\documentclass[10pt,journal,compsoc]{IEEEtran}
\usepackage{amsmath,amsfonts}
\usepackage{algorithmic}
\usepackage{algorithm}
\usepackage{array}
\usepackage[caption=false,font=normalsize,labelfont=sf,textfont=sf]{subfig}
\usepackage{textcomp}
\usepackage{stfloats}
\usepackage{url}
\usepackage{verbatim}
\usepackage{graphicx}
\usepackage{cite}
\usepackage{xcolor}
\usepackage{booktabs}
\usepackage{multirow}

\usepackage{adjustbox}
\usepackage[normalem]{ulem}
\usepackage{hyperref}

\definecolor{myblue}{rgb}{0.2, 0.2, 0.9}

\newcommand{\bmH}{\bm H}
\newcommand{\bmQ}{\bm Q}
\newcommand{\bmK}{\bm K}
\newcommand{\bmV}{\bm V}
\newcommand{\bmW}{\bm W}

\usepackage{microtype}
\RequirePackage{amsmath,amsfonts,amssymb,bm,latexsym}
\usepackage{subfig}
\usepackage{float}
\usepackage{multirow}
\usepackage{tikz}
\usetikzlibrary{fadings}
\usetikzlibrary{mindmap,trees}
\usetikzlibrary{arrows,automata,shapes,positioning,shadows,trees}
\usetikzlibrary{shapes,snakes,shadows}
\usetikzlibrary{shapes.arrows}
\usetikzlibrary{calc,shapes, positioning}
\usetikzlibrary{decorations.text}
\usetikzlibrary{matrix,chains,positioning,decorations.pathreplacing,arrows}
\usetikzlibrary{bayesnet}
\usepackage[edges]{forest}
\usetikzlibrary{shadows.blur}
\usetikzlibrary{shapes.geometric}

\usepackage{pgfplots}
\usepackage{pgfplotstable}
\usepackage{pgfmath,pgffor}
\usepgfplotslibrary{colorbrewer}
\pgfplotsset{compat = 1.14, cycle list/Set1-8}
\usetikzlibrary{pgfplots.statistics, pgfplots.colorbrewer}
\pgfplotsset{compat=1.8}

\NewDocumentCommand{\heng}
{ mO{} }{\textcolor{red}{\textsuperscript{\textit{Heng}}\textsf{\textbf{\small[#1]}}}}

\begin{document}

\title{Large Language Models on Graphs: A Comprehensive Survey}


\author{Bowen Jin*,
        Gang Liu*,
        Chi Han*,
        Meng Jiang,
        Heng Ji,
         Jiawei Han

\IEEEcompsocitemizethanks{\IEEEcompsocthanksitem * The first three authors contributed equally to this work.

\IEEEcompsocthanksitem Bowen Jin, Chi Han, Heng Ji, Jiawei Han: University of Illinois at Urbana-Champaign.  \{bowenj4, chihan3, hengji, hanj\}@illinois.edu

\IEEEcompsocthanksitem Gang Liu, Meng Jiang: University of Notre Dame. \{gliu7, mjiang2@\}@nd.edu

}
}


\markboth{Transactions on Knowledge and Data Engineering (TKDE) 2024.}%
{Shell \MakeLowercase{\textit{et al.}}: A Sample Article Using IEEEtran.cls for IEEE Journals}


\maketitle

\begin{abstract}
Large language models (LLMs), such as GPT4 and LLaMA, are creating significant advancements in natural language processing, due to their strong text encoding/decoding ability and newly found emergent capability (\textit{e.g.}, reasoning).
While LLMs are mainly designed to process pure texts, there are many real-world scenarios where text data is associated with rich structure information in the form of graphs (\textit{e.g.}, academic networks, and e-commerce networks) or scenarios where graph data is paired with rich textual information (\textit{e.g.}, molecules with descriptions).
Besides, although LLMs have shown their pure text-based reasoning ability, it is underexplored whether such ability can be generalized to graphs (\textit{i.e.}, graph-based reasoning).
In this paper, we provide a systematic review of scenarios and techniques related to large language models on graphs.
We first summarize potential scenarios of adopting LLMs on graphs into three categories, namely pure graphs, text-attributed graphs, and text-paired graphs.
We then discuss detailed techniques for utilizing LLMs on graphs, including LLM as Predictor, LLM as Encoder, and LLM as Aligner, and compare the advantages and disadvantages of different schools of models.
Furthermore, we discuss the real-world applications of such methods and summarize open-source codes and benchmark datasets.
Finally, we conclude with potential future research directions in this fast-growing field.
\textbf{The related source can be found at \url{https://github.com/PeterGriffinJin/Awesome-Language-Model-on-Graphs}}. 
\end{abstract}


\vspace{-0.2in}
\section{Introduction}

\begin{figure}
\centering
\includegraphics[width=0.48\textwidth]{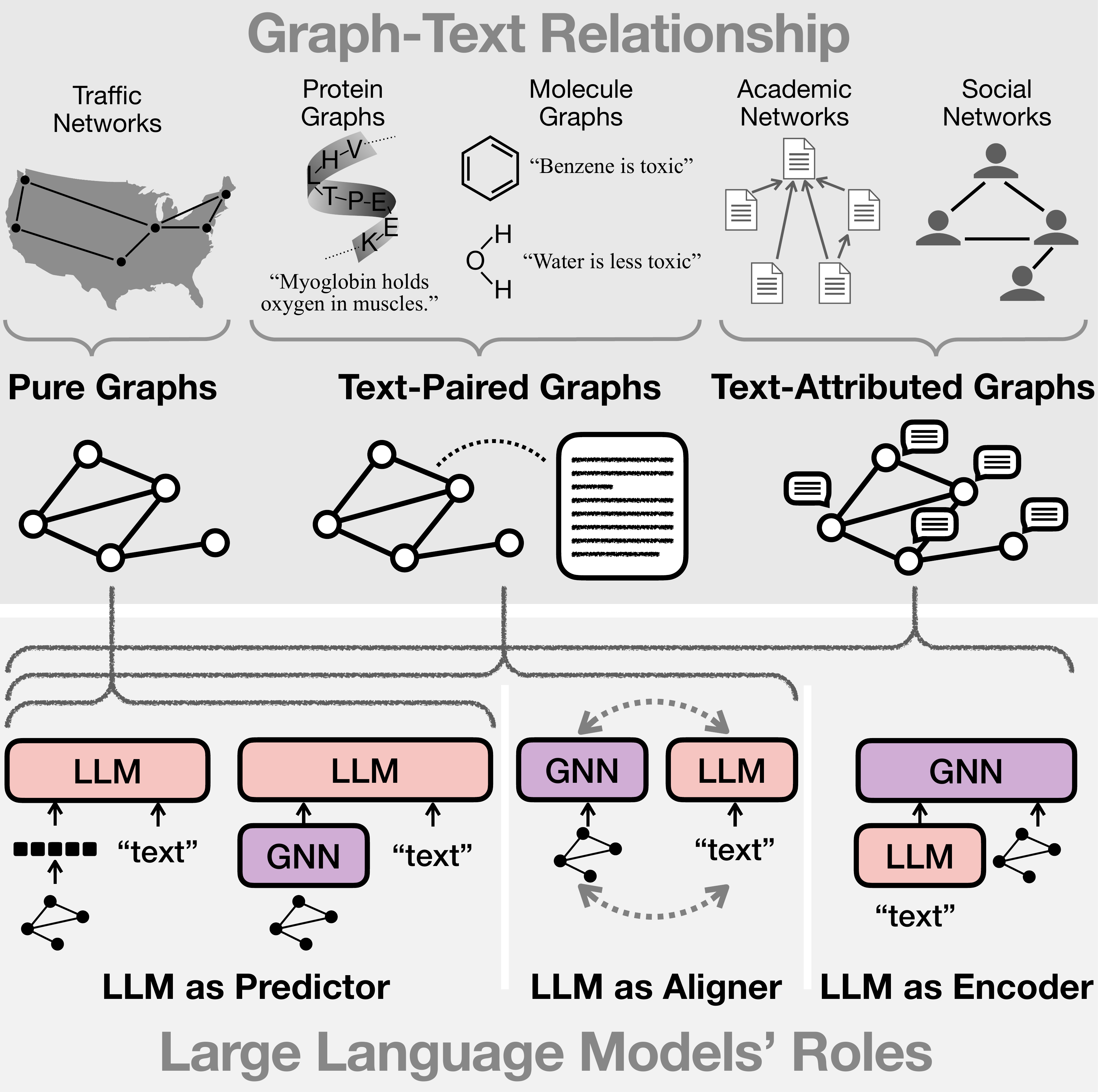}
\caption{According to the relationship between graph and text, we categorize three LLM on graph scenarios. Depending on the role of LLM, we summarize three LLM-on-graph techniques. ``LLM as Predictor'' is where LLMs are responsible for predicting the final answer. ``LLM as Aligner'' will align the inputs-output pairs with those of GNNs. ``LLM as Encoder'' refers to using LLMs to encode and obtain feature vectors. }\label{fig::overview}
\vspace{-0.2in}
\end{figure}

\IEEEPARstart{L}{arge} language models (LLMs) (\textit{e.g.}, BERT \cite{devlin2018bert}, T5 \cite{raffel2020exploring}, LLaMA \cite{touvron2023llama}) which represents a direction of ever-increasing models' sizes pre-trained on larger corpora, have demonstrated powerful capabilities in solving natural language processing (NLP) tasks, including question answering \cite{yang2019end}, text generation \cite{liu2019text} and document understanding \cite{wang2018glue}.
There are no clear and static thresholds regarding the model sizes. Early LLMs (\textit{e.g.}, BERT \cite{devlin2018bert}, RoBERTa \cite{liu2019roberta}) adopt an encoder-only architecture and show capabilities in text representation learning \cite{reimers2019sentence} and natural language understanding \cite{wang2018glue}.
In recent years, more focus has been given to larger decoder-only architectures \cite{touvron2023llama} or encoder-decoder architectures \cite{raffel2020exploring}.
As the model size scales up, such LLMs have also shown reasoning ability and even more advanced emergent ability \cite{wei2022emergent}, exposing a strong potential for Artificial General Intelligence (AGI).

While LLMs are extensively applied to process pure texts, there is an increasing number of applications where the text data are associated with structure information which are represented in the form of graphs.
As presented in Fig.~\ref{fig::overview}, in academic networks, papers (with title and description) and authors (with profile text), are interconnected with authorship relationships. 
Understanding both the author/paper's text information and author-paper structure information on such graphs can contribute to advanced author/paper modeling and accurate recommendations for collaboration;
In the scientific domain, molecules are represented as graphs and are often paired with text that describes their basic properties (\textit{e.g.}, mass and weight). 
Joint modeling of both the molecule structure (graph) and the associated rich knowledge (text) is important for deeper molecule understanding.
Since LLMs are mainly proposed for modeling texts that lie in a sequential fashion, those scenarios mentioned above pose new challenges on how to enable LLMs to encode the structure information on graphs.
In addition, since LLMs have demonstrated their superb text-based reasoning ability, it is promising to explore whether they have the potential to address fundamental graph reasoning problems on pure graphs.
These graph reasoning tasks include inferring connectivity \cite{nagamochi2008algorithmic}, shortest path \cite{goldberg2005computing}, subgraph matching \cite{sun2012efficient}, and logical rule induction \cite{liu2021neural}. 

Recently, there has been an increasing interest~\cite{chen2023llmgraph} in extending LLMs for graph-based applications (summarized in Fig.~\ref{fig::overview}).
According to the relationship between graph and text presented in Fig.~\ref{fig::overview}, the application scenarios can be categorized into pure graphs, text-attributed graphs (nodes/edges are associated with texts), and text-paired graphs.
Depending on the role of LLMs and their interaction with graph neural networks (GNNs), the LLM on graphs techniques can be classified into treating LLMs as the final component for prediction (LLM as Predictor), treating LLMs as the feature extractor for GNNs (LLM as Encoder), and align the latent space of LLMs with GNNs (LLM as Aligner).

There are a limited number of existing surveys exploring the intersection between LLMs and graphs. 
Related to deep learning on graphs,
Liu et al. \cite{graphfoundationsurvey} discuss pretrained foundation models on graphs, including their backbone architectures, pretraining methods, and adaptation techniques.
Pan et al. \cite{pan2023unifying} review the connection between LLMs and knowledge graphs (KGs) especially on how KGs can enhance LLMs training and inference, and how LLMs can facilitate KG construction and reasoning.
Mao et al. \cite{mao2024advancing} and Li et al. \cite{li2023survey} review LLM on graphs focusing on techniques rather than applications.
In summary, existing surveys either focus more on GNNs rather than LLMs or fail to provide a systematic perspective on their applications in various graph scenarios as in Fig.~\ref{fig::overview}.
Our paper provides a comprehensive review of the LLMs on graphs for broader researchers from diverse backgrounds besides the computer science and machine learning community who want to enter this rapidly developing field.

\textbf{Our Contributions.} The notable contributions of our paper are summarized as follows:

\begin{itemize}
    \item \textbf{Categorization of Graph Scenarios.} We systematically summarize the graph scenarios where language models can be adopted into: pure graphs, text-attributed graphs, and text-paired graphs.
    \item \textbf{Systematic Review of Techniques.} We provide the most comprehensive overview of language models on graph techniques. For different graph scenarios, we summarize the representative models, provide detailed illustrations of each of them, and make necessary comparisons.
    \item \textbf{Abundant Resources.} We collect abundant resources on language models on graphs, including benchmark datasets, open-source codebases, and practical applications. 
    \item \textbf{Future Directions.} We delve into the foundational principles of language models on graphs and propose six prospective avenues for future exploration.
\end{itemize}

\textbf{Organization of Survey.}
The rest of this survey is organized as follows.
Section \ref{sec:background} introduces the background of LLMs and GNNs, lists commonly used notations, and defines related concepts.
Section \ref{sec:categorization} categorizes graph scenarios where LLMs can be adopted and summarizes LLMs on graph techniques.
Section \ref{sec:graph-reasoning}-\ref{sec:graph-text} provides a detailed illustration of LLM methodologies for different graph scenarios.
Section \ref{sec:application} delivers available datasets, open-source codebases, and a collection of applications across various domains.
Section \ref{sec:future-direction} introduces some potential future directions.
Section \ref{sec:conclusion} summarizes the paper.

\vspace{-0.2in}
\section{Definitions \& Background}\label{sec:background}

\begin{table}[t]
\caption{Notations of Concepts.}\label{tab:notations}
\vspace{-0.15in}
\centering
\begin{tabular} {llp{6.5cm}} \toprule
\textbf{Notations}& \textbf{Descriptions} \\ \midrule
$|\cdot|$ & The length of a set. \\ \hline
$[\mathbf{A},\mathbf{B}]$ & The concatenation of $\mathbf{A}$ and $\mathbf{B}$. \\ \hline
$\Vert$ & Concatenate operation. \\ \hline
$\mathcal{G}$& A graph. \\ \hline
$\mathcal{V}$& The set of nodes in a graph.\\ \hline
$v$ & A node $v\in \mathcal{V}$. \\ \hline
$\mathcal{E}$& The set of edges in a graph.\\ \hline
$e$ & An edge $e\in \mathcal{E}$.\\ \hline
$\mathcal{G}_v$& The ego graph associated with $v$ in $\mathcal{G}$. \\ \hline
$N(v)$ & The neighbors of a node $v$. \\ \hline
$M$ & A meta-path or a meta-graph.\\ \hline
\multirow{2}{*}{$N_M(v)$} & The nodes which are reachable from  \\
& node $v$ with meta-path or meta-graph $M$. \\ \hline
$\mathcal{D}$ & The text set. \\ \hline
$s \in \mathcal{S}$ & The text token in a text sentence $\mathcal{S}$. \\ \hline
$d_{v_i}$ & The text associated with the node $v_i$. \\ \hline
$d_{e_{ij}}$ & The text associated with the edge $e_{ij}$. \\ \hline
$d_{\mathcal{G}}$ & The text associated with the graph $\mathcal{G}$. \\ \hline
$n$ & The number of nodes, $n = |V|$. \\ \hline
$b$ & The dimension of a node hidden state. \\ \hline
$\mathbf{x}_{v_i} \in \mathbf{R}^d$ & The initial feature vector of the node $v_i$. \\ \hline
$\mathbf{H}_v \in \mathbf{R}^{n \times b}$ & The node hidden feature matrix. \\ \hline
$\mathbf{h}_{v_i} \in \mathbf{R}^{b}$ & The hidden representation of node $v_i$. \\ \hline
{$\mathbf{h}_{\mathcal{G}} \in \mathbf{R}^{b}$} & The hidden representation of a graph $\mathcal{G}$. \\ \hline
{$\mathbf{h}_{d_v} \in \mathbf{R}^{b}$} & The representation of text $d_v$. \\ \hline
{$\mathbf{H}_{d_v} \in \mathbf{R}^{|d_v|\times b}$} & The hidden states of tokens in $d_v$. \\ \hline	
$\mathbf{W},\mathbf{\Theta},w,\theta$ & Learnable model parameters. \\  \hline
$\text{LLM}(\cdot)$ & Large Language model. \\ \hline
$\text{GNN}(\cdot)$ & Graph neural network. \\
\bottomrule
\end{tabular}
\vspace{-0.2in}
\end{table}

\subsection{Definitions}

We provide definitions of various types of graphs and introduce the notations (as shown in Table \ref{tab:notations}) in this section.

\textit{Definition 1 (Graph):} A graph can be defined as $\mathcal{G}=(\mathcal{V}, \mathcal{E})$. Here $\mathcal{V}$ signifies the set of nodes, while $E$ denotes the set of edges. A specific node can be represented by $v_i\in \mathcal{V}$, and an edge directed from node $v_j$ to $v_i$ can be expressed as $e_{ij}=(v_i,v_j)\in \mathcal{E}$. The set of nodes adjacent to a particular node $v$ is articulated as $N(v)=\{u\in \mathcal{V}|(v,u)\in \mathcal{E}\}$.

\textit{Definition 2 (Graph with node-level textual information):}
This type of graph can be denoted as $\mathcal{G}=(\mathcal{V}, \mathcal{E}, \mathcal{D})$, where $\mathcal{V}$, $\mathcal{E}$ and $\mathcal{D}$ are node set, edge set, and text set, respectively. Each $v_i\in \mathcal{V}$ is associated with some textual information $d_{v_i}\in\mathcal{D}$.
For instance, in an academic citation network, one can interpret $v\in\mathcal{V}$ as the scholarly articles, $e\in\mathcal{E}$ as the citation links between them, and $d\in\mathcal{D}$ as the textual content of these articles.
A graph with node-level textual information is also called a \textbf{text-attributed graph} \cite{jin2023patton}, a text-rich graph \cite{zhao2022learning}, or a textual graph \cite{yang2021graphformers}.

\textit{Definition 3 (Graph with edge-level textual information):}
This type of graph can be denoted as $\mathcal{G}=(\mathcal{V}, \mathcal{E}, \mathcal{D})$. Each $e_{ij}\in \mathcal{E}$ is associated with some textual information $d_{e_{ij}}\in\mathcal{D}$.
For example, in a social network, one can interpret $v\in\mathcal{V}$ as the users, $e\in\mathcal{E}$ as the interaction between the users, and $d\in\mathcal{D}$ as the textual content of the messages sent between the users.
Such a graph is also called a textual-edge network \cite{jin2023edgeformers}.

\textit{Definition 4 (Graph with graph-level textual information):}
This type of graph can be denoted as the pair $(\mathcal{G}, d_{\mathcal{G}})$, where $\mathcal{G}=(\mathcal{V}, \mathcal{E})$. $\mathcal{V}$ and $\mathcal{E}$ are node set and edge set. $d_{\mathcal{G}}$ is the text set paired to the graph $\mathcal{G}$.
For instance, in a molecular graph $\mathcal{G}$, $v\in\mathcal{V}$ denotes an atom, $e\in\mathcal{E}$ represents the strong attractive forces or chemical bonds that hold molecules together, and $d_\mathcal{G}$ represents the textual description of the molecule. 
We note that texts may also be associated with subgraph-level concepts and then paired with the entire graph.
Such a graph is also called a \textbf{text-paired graph}.

\vspace{-0.1in}
\subsection{Background}\label{subsec:background}

\textbf{(Large) Language Models.} Language Models (LMs), or language modeling, is an area in the field of natural language processing (NLP) on understanding and generation from text distributions. In recent years, large language models (LLMs) have demonstrated impressive capabilities in tasks such as machine translation, text summarization, reasoning, and question answering \cite{weiemergent, wei2021finetuned, kojimalarge, weichain, brown2020language, radford2019language,yu2022mokge, zhao2023survey}.

Language models have evolved significantly over time.
BERT \cite{devlin2018bert} marks significant progress in language modeling and representation. BERT models the conditional probability of a word given its bidirectional context, also named masked language modeling (MLM) objective
:
\begin{equation}
    \mathbb{E}_{\mathcal{S} \sim \mathcal{D}} \left[ \sum_{s_i\in\mathcal{S}} \log p(s_i | s_1, \dots, s_{i-1}, s_{i+1}, \dots, s_{N_\mathcal{S}}) \right],
\end{equation}
where $\mathcal{S}$ is a sentence sampled from the corpus $\mathcal{D}$, $s_i$ is the $i$-th word in the sentence, and $N_\mathcal{S}$ is the length of the sentence.
On the other hand, the objective of causal language modeling or text generation is defined as:
\begin{equation}
    \mathbb{E}_{\mathcal{S} \sim \mathcal{D}} \left[ \sum_{s_i\in\mathcal{S}} \log p(s_i | s_1, \dots, s_{i-1}) \right].
\end{equation}
Following BERT, other masked language models are proposed, such as RoBERTa \cite{liu2019roberta}, ALBERT \cite{lan2019albert}, and ELECTRA \cite{clark2019electra}, with similar architectures and objectives of text representation.
Efforts have been made to combine language models with other modalities such as vision \cite{radford2021learning, alayrac2022flamingo} and biochemical structures \cite{edwards2021text2mol, edwards2022translation, zhao2023gimlet}. 
In this paper, we will discuss its combination with graphs.

The lifecycle of an LLM usually involves some or all the following steps: pretraining, finetuning, and prompting. In pretraining, LLMs are usually trained on a larger corpus with multiple language modeling objectives \cite{devlin2018bert, brown2020language, lewis2019bart}, which aims to endow LLMs with strong language understanding and completion capability. If domain-specific abilities are expected, LLMs are then finetuned with a smaller amount of domain-specific data~\cite{lester2021power, li2021prefix, houlsby2019parameter, hu2021lora, wei2021finetuned, sanh2021multitask}. Human preference optimization methods are sometimes applied after this stage to align outputs better with users' intentions or social values~\cite{griffith2023rlhf, schulman2017ppo, rafailov2024dpo}. Finally, various prompting or prompt engineering techniques can be deployed to boost downstream task performance \cite{wei2022chain, yao2023tree, besta2023graph}. A more comprehensive description can be found in Appendix~\ref{app:training}

We would like to point out that the word ``large'' in LLM is not associated with a clear and static threshold to divide language models.
``Large'' actually refers to a direction in which language models are inevitably evolving, and larger foundational models tend to possess significantly more representation and generalization power. 
Hence, we define LLMs to encompass both medium-scale PLMs, such as BERT, and large-scale LMs, like GPT-4, as suggested by~\cite{pan2023unifying}.

\vspace{0.5em}
\noindent\textbf{Graph Neural Networks \& Graph Transformers.} 
In real-world scenarios, not all the data are sequential like text, many data lies in a more complex non-Euclidean structure, \textit{i.e.}, graphs.
GNN is proposed as a deep-learning architecture for graph data. Primary GNNs including GCN~\cite{kipf2016semi}, GraphSAGE~\cite{hamilton2017inductive} and, GAT~\cite{velivckovic2017graph} are designed for solving node-level tasks. They mainly adopt a propagation-aggregation paradigm to obtain node representations:
\[
\bm{h}_v^{(l)} = \text{AGG}^{(l)}\left(\bm{h}_v^{(l-1)}, \text{PROP}^{(l)}\left(\{\bm{h}_u^{(l-1)} \mid u \in \mathcal{N}(v)\}\right)\right).
\]
When propagation is global ($u \in \mathcal{V}$), the Graph Transformer~\cite{graphormer,gps} with attention-weighted node importance during sum aggregation can be defined. Let $\mathbf{W}_Q, \mathbf{W}_K, \mathbf{W}_V$ be the query, key, and value matrices, respectively, and $k_{\exp}$ denote the similarity between two nodes. Then, we have:
\[
\text{Attn}(\mathbf{h}_v^{(l-1)}) = \sum_{u \in \mathcal{V}} \frac{k_{\exp}(\mathbf{h}^{(l-1)}_v, \mathbf{h}^{(l-1)}_u)}{\sum_{w \in \mathcal{V}} k_{\exp}(\mathbf{h}^{(l-1)}_v, \mathbf{h}^{(l-1)}_w)} \mathbf{h}^{(l-1)}_u \mathbf{W}_V,
\]
where $k_{\exp}(\mathbf{h}^{(l-1)}_v, \mathbf{h}^{(l)}_w) = \exp \left( \frac{\mathbf{h}^{(l-1)}_v \mathbf{W}_Q \mathbf{h}^{(l-1)}_w \mathbf{W}_K}{\sqrt{d_K}} \right)$.

To solve graph-level tasks, GNN models like GIN~\cite{gin} or Graph Transformers obtain graph representations using a READOUT function: $\bm{h}_{\mathcal{G}} = \text{READOUT}(\{\bm{h}_{v_i} \mid v_i \in \mathcal{G}\})$.
The READOUT functions include mean pooling, max pooling, and so on.
Subsequent work on GNN tackles the issues of over-smoothing~\cite{oversoomth}, over-squashing~\cite{oversquash}, interpretability~\cite{grea}, and bias~\cite{sgir}.
While message-passing-based GNNs excel in structure encoding, researchers aim to enhance their expressiveness with Graph Transformers. These models leverage global multi-head attention mechanisms and integrate graph inductive biases through positional encoding, structural encoding, combining message-passing with attention layers, or improving attention efficiency on large graphs.
Graph Transformers have been proven to be a state-of-the-art solution for many pure graph problems.

\vspace{0.5em}
\noindent\textbf{Language Models vs. Graph Transformers.} 
Modern language models and graph Transformers both use Transformers \cite{vaswani2017attention} as the base model architecture.  This makes the two concepts hard to distinguish, especially when the language models are adopted on graph applications. In this paper, ``Transformers'' typically refers to Transformer language models for simplicity. Here, we provide three points to help distinguish them: 
1) \textit{Tokens} (word token vs. node token): Transformers take a token sequence as inputs. For language models, the tokens are word tokens; while for graph Transformers, the tokens are node tokens. In those cases where tokens include both word tokens and node tokens if the backbone Transformers is pretrained on text corpus (\textit{e.g.}, BERT \cite{devlin2018bert} and LLaMA \cite{touvron2023llama}), we will call it a ``language model''.
2) \textit{Positional Encoding} (sequence vs. graph): language models typically adopt the absolute or relative positional encoding considering the position of the word token in the sequence, while graph Transformers adopt shortest path distance~\cite{graphormer}, random walk distance, the eigenvalues of the graph Laplacian~\cite{gps} to consider the distance of nodes in the graph.
3) \textit{Goal} (text vs. graph): The language models are originally proposed for text encoding and generation; while graph Transformers are proposed for node encoding or graph encoding. 
In those cases where texts are served as nodes/edges on the graph if the backbone Transformers is pretrained on text corpus, we will call it a ``language model''.

\definecolor{connect-line}{RGB}{0,0,0}
\definecolor{middle-color}{RGB}{255,255,255}
\definecolor{leaf-color}{RGB}{255,255,255}
\definecolor{line-color}{RGB}{25,25,112}

\definecolor{black}{RGB}{0,0,0}


\definecolor{pure}{RGB}{112,25,25}
\definecolor{node}{RGB}{25,25,112}
\definecolor{graph}{RGB}{25,112,25}

\tikzstyle{pure-leaf}=[draw=pure,
    rounded corners,minimum height=1em,
    fill=leaf-color!40,text opacity=1, align=center,
    fill opacity=.5,  text=black,align=left,font=\scriptsize,
    inner xsep=3pt,
    inner ysep=1pt,
]
\tikzstyle{pure-middle}=[draw=pure,
    rounded corners,minimum height=1em,
    fill=middle-color!40,text opacity=1, align=center,
    fill opacity=.5,  text=black,align=left,font=\scriptsize,
    inner xsep=3pt,
    inner ysep=1pt,
]
    
\tikzstyle{node-leaf}=[draw=node,
    rounded corners,minimum height=1em,
    fill=leaf-color!40,text opacity=1, align=center,
    fill opacity=.5,  text=black,align=left,font=\scriptsize,
    inner xsep=3pt,
    inner ysep=1pt,
]
\tikzstyle{node-middle}=[draw=node,
    rounded corners,minimum height=1em,
    fill=middle-color!40,text opacity=1, align=center,
    fill opacity=.5,  text=black,align=left,font=\scriptsize,
    inner xsep=3pt,
    inner ysep=1pt,
]

\tikzstyle{graph-leaf}=[draw=graph,
    rounded corners,minimum height=1em,
    fill=leaf-color!40,text opacity=1, align=center,
    fill opacity=.5,  text=black,align=left,font=\scriptsize,
    inner xsep=3pt,
    inner ysep=1pt,
]
\tikzstyle{graph-middle}=[draw=graph,
    rounded corners,minimum height=1em,
    fill=middle-color!40,text opacity=1, align=center,
    fill opacity=.5,  text=black,align=left,font=\scriptsize,
    inner xsep=3pt,
    inner ysep=1pt,
]

\tikzstyle{leaf}=[draw=line-color,
    rounded corners,minimum height=1em,
    fill=leaf-color!40,text opacity=1, align=center,
    fill opacity=.5,  text=black,align=left,font=\scriptsize,
    inner xsep=3pt,
    inner ysep=1pt,
    ]
\tikzstyle{middle}=[draw=line-color,
    rounded corners,minimum height=1em,
    fill=middle-color!40,text opacity=1, align=center,
    fill opacity=.5,  text=black,align=left,font=\scriptsize,
    inner xsep=3pt,
    inner ysep=1pt,
    ]
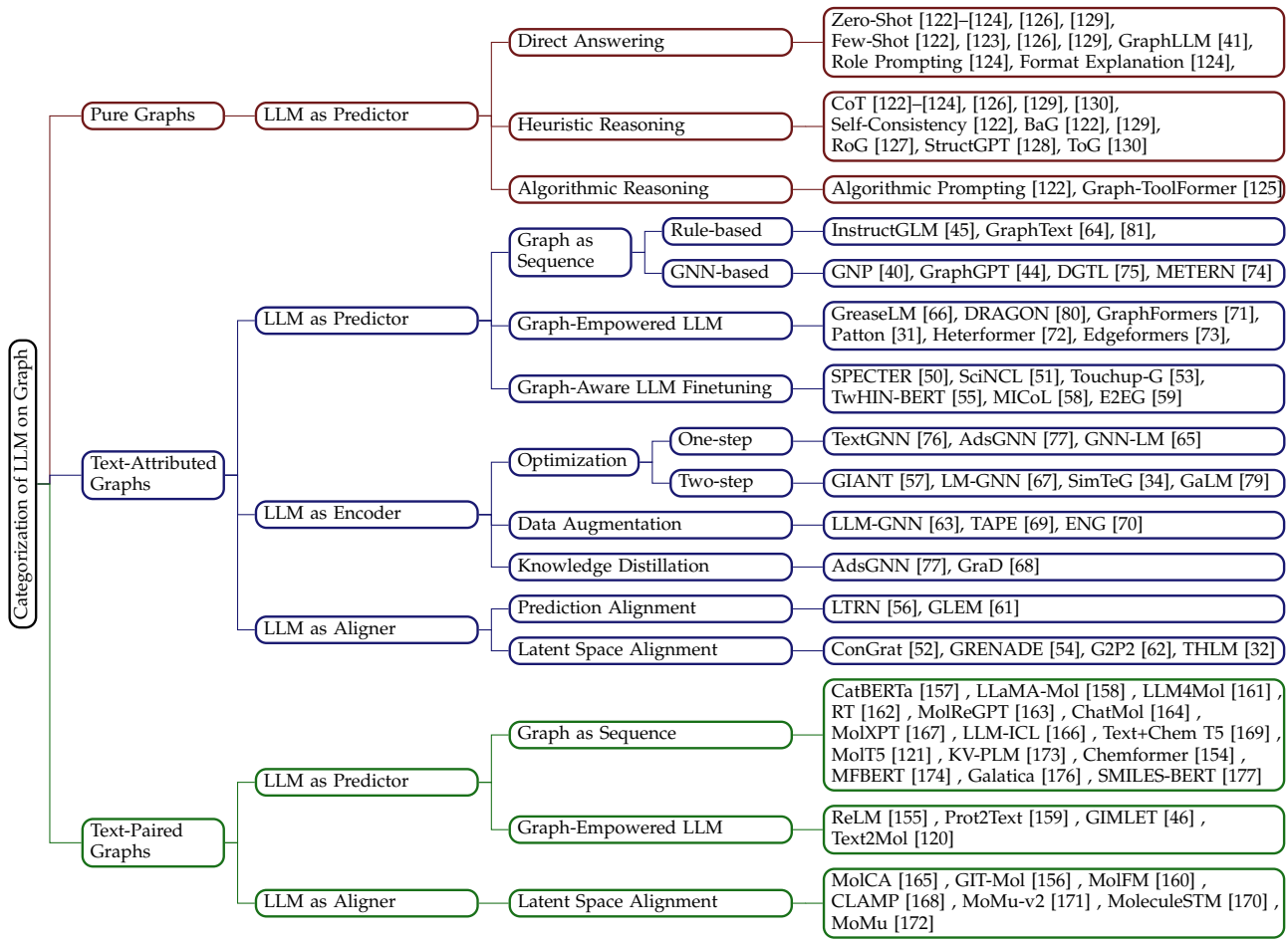
\begin{figure*}[ht]
\centering
\begin{forest}
  for tree={
    forked edges,
    grow=east,
    reversed=true,
    anchor=base west,
    parent anchor=east,
    child anchor=west,
    base=middle,
    font=\scriptsize,
    rectangle,
    line width=0.9pt,
    draw=connect-line,
    rounded corners,align=left,
    minimum width=2em,
    s sep=5pt,
    inner xsep=3pt,
    inner ysep=1pt,
  },
  where level=1{text width=4.5em}{},
  where level=2{text width=6em,font=\scriptsize}{},
  where level=3{font=\scriptsize}{},
  where level=4{font=\scriptsize}{},
  where level=5{font=\scriptsize}{},
  [Categorization of LLM on Graph, black,rotate=90,anchor=north,edge=pure
    [Pure Graphs, pure-middle, edge=pure, text width=5em
        [LLM as Predictor, pure-middle, edge=pure, text width=8.2em
            [Direct Answering, pure-middle, text width=10.6em, edge=pure
                [Zero-Shot\cite{wang2023can, liu2023evaluating, guo2023gpt4graph, zhang2023llm4dyg, fatemi2023talk}{,} \\
                Few-Shot\cite{wang2023can, liu2023evaluating, zhang2023llm4dyg, fatemi2023talk}{,}
                GraphLLM\cite{chai2023graphllm}{,} \\
                Role Prompting\cite{guo2023gpt4graph}{,} Format Explanation\cite{guo2023gpt4graph}{,}, pure-leaf, text width=17.7em, edge=pure]
            ]
            [Heuristic Reasoning, pure-middle, text width=10.6em, edge=pure
                [CoT\cite{wang2023can, liu2023evaluating, guo2023gpt4graph, zhang2023llm4dyg, fatemi2023talk, sun2023think}{,}\\
                Self-Consistency\cite{wang2023can}{,} BaG\cite{wang2023can, fatemi2023talk}{,} \\
                RoG\cite{luo2023reasoning}{,} StructGPT\cite{jiang2023structgpt}{,} ToG\cite{sun2023think}
                , pure-leaf, text width=17.7em, edge=pure]
            ]
            [Algorithmic Reasoning, pure-middle, text width=10.6em, edge=pure
                [Algorithmic Prompting\cite{wang2023can}{,} 
                Graph-ToolFormer\cite{zhang2023graph}, pure-leaf, text width=17.7em, edge=pure]
            ]
        ]
    ]
    [Text-Attributed \\ Graphs, node-middle, edge=node, text width=5em
        [LLM as Predictor, node-middle, edge=node,text width=8.2em
            [Graph as \\ Sequence, node-middle, text width=4.2em, edge=node
                [Rule-based, node-middle, text width=4.5em, edge=node
                    [InstructGLM \cite{ye2023natural}{,}  GraphText \cite{zhao2023graphtext}{,} \cite{huang2023can}{,}, node-leaf, text width=17.7em, edge=node]
                ]
                [GNN-based, node-middle, text width=4.5em, edge=node
                    [GNP \cite{tian2023graph}{,} GraphGPT \cite{tang2023graphgpt}{,} DGTL \cite{qin2023disentangled}{,} METERN \cite{jin2023learning} , node-leaf, text width=17.7em, edge=node]
                ]
            ]
            [Graph-Empowered LLM, node-middle, text width=10.6em, edge=node
                [GreaseLM \cite{zhang2022greaselm}{,} DRAGON \cite{yasunaga2022deep}{,} GraphFormers \cite{yang2021graphformers}{,}\\ Patton \cite{jin2023patton}{,} Heterformer \cite{jin2023heterformer}{,} Edgeformers \cite{jin2023edgeformers}{,}, node-leaf, text width=17.7em, edge=node]
            ]
            [Graph-Aware LLM Finetuning, node-middle, text width=10.6em, edge=node
                [SPECTER \cite{cohan2020specter}{,} SciNCL \cite{ostendorff2022neighborhood}{,} Touchup-G \cite{zhu2023touchup}{,} \\ TwHIN-BERT \cite{zhang2023twhin}{,} MICoL \cite{zhang2022metadata}{,} E2EG \cite{dinh2022e2eg} , node-leaf, text width=17.7em, edge=node]
            ]
        ]
        [LLM as Encoder, node-middle, edge=node, text width=8.2em
            [Optimization, node-middle, text width=4.5em, edge=node
                [One-step, node-middle, text width=4.2em, edge=node
                [TextGNN \cite{zhu2021textgnn}{,} AdsGNN \cite{li2021adsgnn}{,} GNN-LM \cite{meng2021gnn}, node-leaf, text width=17.7em, edge=node]
                ]
                [Two-step, node-middle, text width=4.2em, edge=node
                    [GIANT \cite{chien2021node}{,} LM-GNN \cite{ioannidis2022efficient}{,} SimTeG \cite{duan2023simteg}{,} GaLM \cite{xie2023graph}, node-leaf, text width=17.7em, edge=node]
                ]
            ]
            [Data Augmentation, node-middle, text width=10.6em, edge=node
                [LLM-GNN \cite{chen2023label}{,} TAPE \cite{he2023explanations}{,} ENG \cite{yu2023empower}, node-leaf, text width=17.7em, edge=node]
            ]
            [Knowledge Distillation, node-middle, text width=10.6em, edge=node
                [AdsGNN \cite{li2021adsgnn}{,} GraD \cite{mavromatis2023train}, node-leaf, text width=17.7em, edge=node]
            ]
        ]
        [LLM as Aligner, node-middle, edge=node, text width=8.2em
            [Prediction Alignment, node-middle, text width=10.6em, edge=node
                [LTRN \cite{zhang2021minimally}{,} GLEM \cite{zhao2022learning}, node-leaf, text width=17.7em, edge=node]
            ]
            [Latent Space Alignment, node-middle, text width=10.6em, edge=node
                [ConGrat \cite{brannon2023congrat}{,}  GRENADE \cite{li2023grenade}{,} G2P2 \cite{wen2023augmenting}{,} THLM \cite{zou2023pretraining} ,node-leaf, text width= 17.7em, edge=node]
            ]
        ]
    ]
    [Text-Paired \\ Graphs, graph-middle, edge=graph, text width=5em
        [LLM as Predictor, graph-middle, edge=graph, text width=8.2em
            [Graph as Sequence, graph-middle, text width=10.6em, edge=graph
                [CatBERTa~\cite{catberta} {,} LLaMA-Mol~\cite{llamamol} {,} LLM4Mol~\cite{llm4mol} {,} \\ RT~\cite{regt} {,} MolReGPT~\cite{molregpt} {,} ChatMol~\cite{chatmol} {,} \\  MolXPT~\cite{molxpt} {,} LLM-ICL~\cite{llmicl} {,} Text+Chem T5~\cite{tct} {,} \\ MolT5~\cite{edwards2022translation} {,} KV-PLM~\cite{kvplm} {,} Chemformer~\cite{chemformer} {,} \\ MFBERT~\cite{mfbert} {,} Galatica~\cite{galatica} {,} SMILES-BERT~\cite{smilesbert} , graph-leaf, text width=17.7em, edge=graph]
            ]
            [Graph-Empowered LLM, graph-middle, text width=10.6em, edge=graph
                [ReLM~\cite{relm} {,} Prot2Text~\cite{prot2text} {,} GIMLET~\cite{zhao2023gimlet} {,} \\ Text2Mol~\cite{edwards2021text2mol} ,graph-leaf, text width= 17.7em, edge=graph]
            ]
        ]
        [LLM as Aligner, graph-middle, edge=graph, text width=8.2em
            [Latent Space Alignment, graph-middle, text width=10.6em, edge=graph
                [MolCA~\cite{molca} {,} GIT-Mol~\cite{gitmol} {,} MolFM~\cite{molfm} {,} \\ CLAMP~\cite{clamp} {,} MoMu-v2~\cite{momuv2} {,} MoleculeSTM~\cite{moleculestm} {,} \\ MoMu~\cite{momu} ,graph-leaf, text width= 17.7em, edge=graph]
            ]
        ]
    ]
  ]  
\end{forest}
\vspace{-0.15in}
\caption{A taxonomy of LLM on graph scenarios and techniques with representative examples.}
\vspace{-0.2in}
\label{fig:taxonomy-techniques}
\end{figure*}

\vspace{-0.1in}
\section{Categorization and Framework}\label{sec:categorization} 
In this section, we first introduce our categorization of graph scenarios where language models can be adopted. 
Then we discuss the categorization of LLM on graph techniques.
Finally, we summarize the training \& inference framework for language models on graphs.

\vspace{-0.1in}
\subsection{Categorization of Graph Scenarios with LLMs}

\noindent\textbf{Pure Graphs without Textual Information} are graphs with no text information or no semantically rich text information.
Examples include traffic graphs and power transmission graphs.
Those graphs often serve as context to test the graph reasoning ability of large language models (solve graph theory problems) or serve as knowledge sources to enhance the large language models (alleviate hallucination).

\noindent\textbf{Text-Attributed Graphs} refer to graphs where nodes or edges are associated with semantically rich text information.
They are also called text-rich networks \cite{jin2023patton}, textual graphs \cite{yang2021graphformers} or textual-edge networks \cite{jin2023edgeformers}.
Examples include academic networks, e-commerce networks, social networks, and legal case networks.
On these graphs, researchers are interested in learning representations for nodes or edges with both textual and structure information \cite{yang2021graphformers}\cite{jin2023edgeformers}.

\noindent\textbf{Text-Paired Graphs} have textual descriptions defined for the entire graph structure. For example, graphs like molecules may be paired with captions or textual features.
While the graph structure significantly contributes to molecular properties, text descriptions can complement our understanding of molecules.
The graph scenarios can be found in Fig. \ref{fig::overview}.

\vspace{-0.1in}
\subsection{Categorization of LLMs on Graph Techniques}\label{sec:categorizatoin-technique}

According to the roles of LLMs and what are the final components for solving graph-related problems,
we classify LLM on graph techniques into three main categories:

\noindent\textbf{LLM as Predictor}.
This category of methods serves LLM as the final component to output representations or predictions. It can be enhanced with GNNs and can be classified depending on how the graph information is injected into LLM: 
1) \textit{Graph as Sequence}: This type of method makes no changes to the LLM architecture, but makes it be aware of graph structure by taking a ``graph token sequence'' as input. The ``graph token sequence'' can be natural language descriptions for a graph or hidden representations outputted by graph encoders.
2) \textit{Graph-Empowered LLM}: This type of method modifies the architecture of the LLM base model (\textit{i.e.}, Transformers) and enables it to conduct joint text and graph encoding inside their architecture.
3) \textit{Graph-Aware LLM Finetuning}: This type of method makes no changes to the input of the LLMs or LLM architectures, but only fine-tunes the LLMs with supervision from the graph.

\noindent\textbf{LLM as Encoder.}
This method is mostly utilized for graphs where nodes or edges are associated with text information (solving node-level or edge-level tasks). GNNs are the final components and we adopt LLM as the initial text encoder. To be specific, LLMs are first utilized to encode the text associated with the nodes/edges. The outputted feature vectors by LLMs then serve as input embeddings for GNNs for graph structure encoding. The output embeddings from the GNNs are adopted as final node/edge representations for downstream tasks.
However, these methods suffer from convergence issues, sparse data issues, and inefficient issues, where we summarize solutions from optimization, data augmentation, and knowledge distillation perspectives.

\noindent\textbf{LLM as Aligner.}
This category of methods adopts LLMs as text-encoding components and aligns them with GNNs which serve as graph structure encoding components. 
LLMs and GNNs are adopted together as the final components for task solving.
To be specific, the alignment between LLMs and GNNs can be categorized into 1) \textit{Prediction Alignment} where the generated pseudo labels from one modality are utilized for training on the other modality in an iterative learning fashion and 2) \textit{Latent Space Alignment} where contrastive learning is adopted to align text embeddings generated by LLMs and graph embeddings generated by GNNs.

In the following sections, we will follow our categorization in Section \ref{sec:categorization} and discuss detailed methodologies for each graph scenario.

\vspace{-0.15in}
\section{Pure Graphs}\label{sec:graph-reasoning}

The study of pure graphs in graph theory is essential for understanding the introduction of LLMs into graph-related reasoning problems. Pure graphs are a universal representation format used to address a wide range of algorithmic problems in computer science. Many graph-based concepts, such as shortest paths, specific sub-graphs, and flow networks, are strongly connected to real-world applications~\cite{chen1996developing, iqbal2018airline, jiang2022scheduling,wangmdqa}. Therefore, reasoning based on pure graphs is crucial for providing theoretical solutions and insights for real-world applications.

Nevertheless, many reasoning tasks require a computation capacity beyond traditional GNNs.
GNNs are typically designed to carry out a bounded number of operations given a graph size. In contrast, graph reasoning problems can require up to indefinite complexity depending on the task's nature.
On the other hand, LLMs demonstrate excellent emergent reasoning ability~\cite{weiemergent, kojimalarge, wei2022chain} recently.
This is partially due to their autoregressive mechanism, which enables computing indefinite sequences of intermediate steps with careful prompting or training~\cite{wei2022chain, yao2023tree}.

The following subsections discuss the attempts to incorporate LLMs into pure graph reasoning problems. We will also discuss the corresponding challenges, limitations, and findings.
Table~\ref{tab:reasoning-methods} in the Appendix lists a categorization of these efforts.
Usually, input graphs are serialized as part of the input sequence, either by verbalizing the graph structure~\cite{wang2023can, liu2023evaluating, guo2023gpt4graph, zhang2023llm4dyg, luo2023reasoning, jiang2023structgpt, fatemi2023talk, sun2023think}
or by encoding the graph structure into implicit feature sequences~\cite{chai2023graphllm}.
The studied reasoning problems range from simpler ones like connectivity, shortest paths, and cycle detection to harder ones like maximum flow and Hamiltonian pathfinding (an NP-complete problem). A comprehensive list of the studied problems is listed in Appendix Table~\ref{tab:reasoning-problems}. Note that we only list representative problems here. This table does not include more domain-specific problems, such as the spatial-temporal reasoning problems in~\cite{zhang2023llm4dyg}. We first briefly describe the approaches to formatting the graph inputs to be fed to LLMs.

\noindent{\textbf{Plainly Verbalizing Graphs}.}
Verbalizing the graph structure in natural language is the most straightforward way of representing graphs.
Representative approaches include describing the edge and adjacency lists, widely studied in~\cite{wang2023can, liu2023evaluating, zhang2023llm4dyg, fatemi2023talk}.
For example, for a triangle graph with three nodes, the edge list can be written as ``\textit{[(0, 1), (1, 2), (2, 0)]}'', which means node 0 is connected to node 1, node 1 is connected to node 2, node 2 is connected to node 0.
It can also be written in natural language such as ``\textit{There is an edge between node 0 and node 1, an edge between node 1 and node 2, and an edge between node 2 and node 0.}''
On the other hand, we can describe the adjacency list from the nodes' perspective.
For example, for the same triangle graph, the adjacency list can be written as ``\textit{Node 0 is connected to node 1 and node 2. Node 1 is connected to node 0 and node 2. Node 2 is connected to node 0 and node 1.}''

\noindent{\textbf{Paraphrasing Graphs}.}
The verbalized graphs can be lengthy, unstructured, and complicated to read, even for humans, so they might not be the best input format for LLMs to infer the answers.
To this end, researchers also attempt to paraphrase the graph structure into more natural or concise sentences.
\cite{guo2023gpt4graph} find that by prompting LLMs to generate a format explanation of the raw graph inputs for itself (\textit{Format-Explanation}) or to pretend to play a role in a natural task (\textit{Role Prompting}), the performance on some problems can be improved but not systematically.
\cite{fatemi2023talk} explores the effect of grounding the pure graph in a real-world scenario, such as social networks, friendship graphs, or co-authorship graphs.
In such graphs, nodes are described as people, and edges are relationships between people.

\noindent{\textbf{Encoding Graphs Into Implicit Feature Sequences}.}
Finally, researchers also attempt to encode the graph structure into implicit feature sequences as part of the input sequence~\cite{chai2023graphllm}.
Unlike the previous verbalizing approaches, this usually involves training a graph encoder to encode the graph structure into a sequence of features and fine-tuning the LLMs to adapt to the new input format.

\vspace{-0.12in}
\subsection{Direct Answering}
Although graph-based reasoning problems usually involve complex computation, researchers still attempt to let language models directly generate answers from the serialized input graphs as a starting point, partially because of the simplicity of the approach and partially in awe of other emergent abilities of LLMs.
Although various attempts have been made to optimize how graphs are presented in the input sequence discussed in the sections above, bounded by the finite sequence length and computational operations, this approach has a fundamental limitation to solving complex reasoning problems such as NP-complete ones.
Unsurprisingly, most studies find that LLMs possess preliminary graph understanding ability, but the performance is less satisfactory on more complex problems or larger graphs~\cite{wang2023can, liu2023evaluating, guo2023gpt4graph, chai2023graphllm, zhang2023llm4dyg, fatemi2023talk} where reasoning is necessary.

On plainly verbalized graphs, one can prompt LLMs to answer questions either in zero-shot or few-shot (in-context learning) settings. The former asks questions directly given the graph structure, while the latter asks questions about the graph structure after providing a few examples of questions and answers.
\cite{wang2023can, liu2023evaluating, guo2023gpt4graph} do confirm that LLMs can answer easier questions such as connectivity, neighbor identification, and graph size counting but fail to answer more complex questions such as cycle detection and Hamiltonian pathfinding.
Their results also reveal that providing more examples in the few-shot setting increases the performance, especially on easier problems, although it is still not satisfactory. Results on paraphrased graphs indicate that encoding in real-world scenarios can improve performance on some problems, but it still cannot be done consistently. By encoding graphs into features, \cite{chai2023graphllm} demonstrates drastic performance improvement on problems including substructure counting, maximum triplet sum, shortest path, and bipartite matching. This indicates that fine-tuning LLMs has great fitting power on a specific task distribution.

\vspace{-0.12in}
\subsection{Heuristic Reasoning}
Direct mapping to the output leverages the LLMs' powerful representation power to ``guess'' the answers. Still, it does not fully utilize the LLMs' impressive emergent reasoning ability, which is essential for solving complex reasoning problems.
To this end, attempts have been made to let LLMs perform heuristic reasoning on graphs. This approach encourages LLMs to perform a series of intermediate reasoning steps that might heuristically lead to the correct answer, which resembles a path-finding reasoning schema\cite{li2021future}.

\noindent{\textbf{Reasoning Step by Step}.}
Encouraged by the success of chain-of-thought (CoT) reasoning~\cite{wei2022chain, kojimalarge}, researchers also attempt to let LLMs perform reasoning step by step on graphs.
Chain-of-thought encourages LLMs to roll out a sequence of reasoning steps to solve a problem, similar to how humans solve problems.
Zero-shot CoT is a similar approach that does not require any examples.
These techniques are studied in~\cite{wang2023can, liu2023evaluating, guo2023gpt4graph, fatemi2023talk, zhang2023llm4dyg, chai2023graphllm, sun2023think}.
Results indicate that CoT-style reasoning can improve the performance on simpler problems, such as cycle detection and shortest path detection. Still, the improvement is inconsistent or diminishes on more complex problems, such as Hamiltonian path finding and topological sorting.

\noindent{\textbf{Retrieving Subgraphs as Evidence}.}
Many graph reasoning problems, such as node degree counting and neighborhood detection, only involve reasoning on a subgraph of the whole graph.
Such properties allow researchers to let LLMs retrieve the subgraphs as evidence and perform reasoning on the subgraphs.
Build-a-Graph prompting~\cite{wang2023can} encourages LLMs to reconstruct the relevant graph structures and then perform reasoning on them.
This method demonstrates promising results on problems except for Hamiltonian pathfinding, a notoriously tricky problem requiring reasoning on the whole graph.
Another approach, Context-Summarization~\cite{guo2023gpt4graph}, encourages LLMs to summarize the key nodes, edges, or sub-graphs and perform reasoning.

\noindent{\textbf{Searching on Graphs}.}
This kind of reasoning is related to the search algorithms on graphs, such as breadth-first search (BFS) and depth-first search (DFS)
Although not universally applicable, BFS and DFS are the most intuitive and effective ways to solve some graph reasoning problems.
Numerous explorations have been made to simulate searching-based reasoning, especially on knowledge-graph question answering.
This approach enjoys the advantage of providing interpretable evidence besides the answer.
Reasoning-on-Graphs (RoG)~\cite{luo2023reasoning} is a representative approach that prompts LLMs to generate several relation paths as plans, which are then retrieved from the knowledge graph (KG) and used as evidence to answer the questions.
Another approach is to iteratively retrieve and reason on the subgraphs from KG~\cite{jiang2023structgpt, sun2023think}, simulating a dynamic searching process.
At each step, the LLMs retrieve neighbors of the current nodes and then decide to answer the question or continue the next search step. These methods address the scalability challenge when knowledge from multiple graphs is available.

\vspace{-0.15in}
\subsection{Algorithmic Reasoning}
The previous two approaches are heuristic, which means that the reasoning process accords with human intuition but is not guaranteed to lead to the correct answer.
In contrast, these problems are usually solved by algorithms in computer science.
Therefore, researchers also attempt to let LLMs perform algorithmic reasoning on graphs.
\cite{wang2023can} proposed ``\textit{Algorithmic Prompting}'', which prompts the LLMs to recall the algorithms that are relevant to the questions and then perform reasoning step by step according to the algorithms.
Their results, however, do not show consistent improvement over the heuristic reasoning approach.
A more direct approach, Graph-ToolFormer~\cite{zhang2023graph}, lets LLMs generate API calls as explicit reasoning steps.
These API calls are then executed externally to acquire answers on an external graph.
This approach is suitable for converting real-world tasks into pure graph reasoning problems, and it has demonstrated efficacy in various applications such as knowledge graphs, social networks, and recommendation systems.

\vspace{-0.15in}
\subsection{Discussion}
Despite the extensive research, there has not been a consensus about the best practice in graph representation in LLMs. The eventual solution to this problem should reach a perfect balance between computation efficiency and information completeness, probably drawing inspiration from long-context LLM researches~\cite{han2023lminfinite, su2024roformer}. The above reasoning methods are not mutually exclusive, and future efforts can be made to combine them to achieve better performance. For example, efficiency in algorithmic searching can be improved by prompting language models for better heuristics.

\vspace{-0.15in}
\section{Text-Attributed Graphs}\label{sec:graph-node-text}
Text-attributed graphs exist ubiquitously in the real world, \textit{e.g.}, academic networks, and legal case networks.
Learning on such networks requires the model to encode both the textual information associated with the nodes/edges and the structure information lying inside the input graph.
Depending on the role of LLM, existing works can be categorized into three types: LLM as Predictor, LLM as Encoder, and LLM as Aligner.
We summarize all surveyed methods in Appendix Table~\ref{tab:node-text-papers}.

\vspace{-0.1in}
\subsection{LLM as Predictor} \label{subsec:graph-node-text-LM}
These methods serve the language model as the main model architecture to capture both the text information and graph structure information.
They can be categorized into three types: \textit{Graph as Sequence methods}, \textit{Graph-Empowered LLMs}, and \textit{Graph-Aware LLM finetuning methods}, depending on how structure information in graphs is injected into language models (input vs. architecture vs. loss). 
In the \textit{Graph as Sequence methods}, graphs are converted into sequences that can be understood by language models together with texts from the inputs.
In the \textit{Graph-Empowered LLMs} methods, people modify the architecture of Transformers (which is the base architecture for LLMs) to enable it to encode text and graph structure simultaneously.
In the \textit{Graph-Aware LLM finetuning methods}, LLM is fine-tuned with graph structure supervision and can generate graph-contextualized representations.

\vspace{-0.1in}
\subsubsection{Graph as Sequence}\label{5.graph-as-sequence}
\begin{figure*}
\centering
\includegraphics[width=17cm]{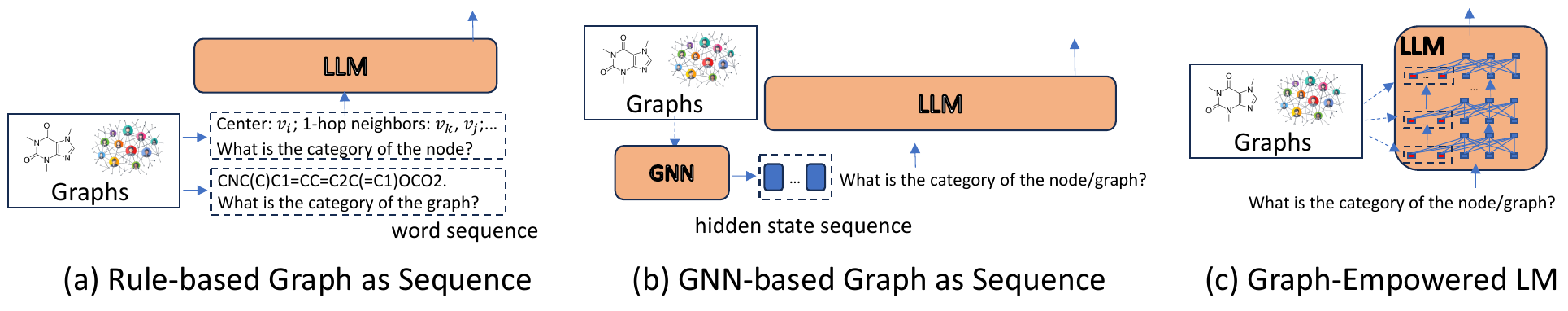}
\vspace{-0.15in}
\caption{The illustration of various LLM as Predictor methods, including (a) Rule-based Graph As Sequence, (b) GNN-based Graph As Sequence, (c) Graph-Empowered LLMs.}\label{fig::lm-centric}
\vspace{-0.2in}
\end{figure*}

In these methods, the graph information is mainly encoded into the LLM from the ``input'' side.
The ego-graphs associated with nodes/edges are serialized into a sequence $\mathbf{H}_{\mathcal{G}_v}$ which can be fed into the LLM together with the texts $d_v$:
\begin{gather}
    \mathbf{H}_{\mathcal{G}_v}= \text{Graph2Seq}(\mathcal{G}_v), \\
    \mathbf{h}_v = \text{LLM}([\mathbf{H}_{\mathcal{G}_v}, d_v]).
\end{gather}
Depending on the choice of $\text{Graph2Seq}(\cdot)$ function, the methods can be further categorized into rule-based methods and GNN-based methods.
The illustration of the categories can be found in Fig. \ref{fig::lm-centric}.

\vspace{0.3em}
\noindent\textbf{Rule-Based: Linearizing Graphs into Text Sequence with Rules.} 
These methods design rules to describe the structure with natural language and adopt a text prompt template as $\text{Graph2Seq}(\cdot)$.
For example, given an ego-graph $\mathcal{G}_{v_i}$ of the paper node $v_i$ connecting to author nodes $v_j$ and $v_k$ and venue nodes $v_t$ and $v_s$, $\mathbf{H}_{\mathcal{G}_{v_i}}=\text{Graph2Seq}(\mathcal{G}_{v_i})=$ ``\textit{The centor paper node is $v_i$. Its author neighbor nodes are $v_j$ and $v_k$ and its venue neighbor nodes are $v_t$ and $v_s$}''.
This is the most straightforward and easiest way (without introducing extra model parameters) to encode graph structures into language models.
Along this line, InstructGLM \cite{ye2023natural} designs templates to describe local ego-graph structure (maximum 3-hop connection) for each node and conduct instruction tuning for node classification and link prediction.
GraphText \cite{zhao2023graphtext} further proposes a syntax tree-based method to transfer structure into text sequence. 
Researchers \cite{huang2023can} also study when and why the linearized structure information on graphs can improve the performance of LLM on node classification and find that the structure information is beneficial when the textual information associated with the node is scarce (in this case, the structure information can provide auxiliary information gain). 

\vspace{0.3em}
\noindent\textbf{GNN-Based: Encoding Graphs into Special Tokens with GNNs.} 
Different from rule-based methods which use natural language prompts to linearize graphs into sequences, GNN-based methods adopt graph encoder models (\textit{i.e.}, GNN) to encode the ego-graph associated with nodes into special token representations which are concatenated with the pure text information into the language model:
\begin{gather}
    \mathbf{H}_{\mathcal{G}_v}= \text{Graph2Seq}(\mathcal{G}_v)=\text{GraphEnc}(\mathcal{G}_v).
\end{gather}
The strength of these methods is they can capture hidden representations of useful structure information with a strong graph encoder, while the challenge is how to fill the gap between graph modality and text modality.
GNP \cite{tian2023graph} adopts a similar philosophy from LLaVA \cite{liu2023visual}, where they utilize GNN to generate graph tokens and then project the graph tokens into the text token space with learnable projection matrices. The projected graph tokens are concatenated with text tokens and fed into the language model.
GraphGPT \cite{tang2023graphgpt} further proposes to train a text-grounded GNN for the projection with a text encoder and contrastive learning.
DGTL \cite{qin2023disentangled} introduces disentangled graph learning, serves graph representations as positional encoding, and adds them to the text sequence.
METERN \cite{jin2023learning} adds learnable relation embeddings to node textual sequences for text-based multiplex representation learning on graphs \cite{park2020unsupervised}.

\vspace{-0.1in}
\subsubsection{Graph-Empowered LLMs}
In these methods, researchers design advanced LLM architecture (\textit{i.e.}, Graph-Empowered LLMs) which can conduct joint text and graph encoding inside their model architecture.
Transformers \cite{vaswani2017attention} serve as the base model for nowadays pretrained LMs \cite{devlin2018bert} and LLMs \cite{kasneci2023chatgpt}.
However, they are designed for natural language (sequence) encoding and do not take non-sequential structure information into consideration.
To this end, Graph-Empowered LLMs are proposed.
They have a shared philosophy of introducing virtual structure tokens $\bmH_{\mathcal{G}_v}$ inside each Transformer layer:
\begin{gather}
    \widetilde{\bmH}_{d_v}^{(l)} = [\bmH_{\mathcal{G}_v}^{(l)}, \bmH_{d_v}^{(l)}]
\end{gather}
where $\bmH_{\mathcal{G}_v}$ can be learnable embeddings or output from graph encoders.
Then the original multi-head attention (MHA) in Transformers is modified into an asymmetric MHA to take the structure tokens into consideration:
\begin{equation}
\small
\begin{split}
    {\rm MHA}_{asy}(\bmH_{d_v}^{(l)}, { \widetilde{\bmH}_{d_v}^{(l)}}) &= \Vert_{u=1}^U \ {\rm head}_u(\bmH_{d_v}^{(l)}, { \widetilde{\bmH}_{d_v}^{(l)}}), \\
    \text{where \ \ } {\rm head}_u(\bmH_{d_v}^{(l)}, { \widetilde{\bmH}_{d_v}^{(l)}}) &= {\rm softmax}\Bigg(\frac{\bmQ_u^{(l)}{\widetilde{\bmK}_u^{(l)\top}}}{\sqrt{d/U}}\Bigg) \cdot { \widetilde{\bmV}_u^{(l)}}, \\
    \bmQ_u^{(l)} = \bmH_{d_v}^{(l)} \bmW_{Q,u}^{(l)}, \ \ { \widetilde{\bmK}_u^{(l)}} &= { \widetilde{\bmH}_{d_v}^{(l)}} \bmW_{K,u}^{(l)}, \ \ { \widetilde{\bmV}_u^{(l)}} = { \widetilde{\bmH}_{d_v}^{(l)}} \bmW_{V,u}^{(l)}.
\end{split}
\label{eqn:mha2}
\end{equation}
With the asymmetric MHA mechanism, the node encoding process of the $(l + 1)$-th layer will be:
\begin{equation}
    \begin{gathered}
        \widetilde{\bm{H}}^{(l)'}_{d_v} = {\rm Normalize}(\bm{H}^{(l)}_{d_v} + {\rm MHA}_{asy}(\widetilde{\bm{H}}^{(l)}_{d_v}, \bm{H}^{(l)}_{d_v})), \\
        \bm{H}^{(l+1)}_{d_v} = {\rm Normalize}(\widetilde{\bm{H}}^{(l)'}_{d_v} + {\rm MLP}(\widetilde{\bm{H}}^{(l)'}_{d_v})).
    \end{gathered}
\end{equation}
Along this line of work, GreaseLM \cite{zhang2022greaselm} proposes to have a language encoding component and a graph encoding component in each layer. These two components interact through a modality-fusion layer (MInt layer), where a special structure token is added to the text Transformer input, and a special node is added to the graph encoding layer.
DRAGON \cite{yasunaga2022deep} further proposes strategies to pretrain GreaseLM with unsupervised signals.
GraphFormers \cite{yang2021graphformers} are designed for node representation learning on homogeneous text-attributed networks where the current layer \texttt{[CLS]} token hidden states of neighboring documents are aggregated and added as a new token on the current layer center node text encoding. 
Patton \cite{jin2023patton} proposes to pretrain GraphFormers with two novel strategies: network-contextualized masked language modeling and masked node prediction.
Heterformer \cite{jin2023heterformer} introduces virtual neighbor tokens for text-rich neighbors and textless neighbors which are concatenated with the original text tokens and fed into each Transformer layer.
Edgeformers \cite{jin2023edgeformers} are proposed for representation learning on textual-edge networks where edges are associated with rich textual information.
When conducting edge encoding, virtual node tokens will be concatenated onto the original edge text tokens for joint encoding.

\subsubsection{Graph-Aware LLM Finetuning}
In these methods, the graph information is mainly injected into the LLM by ``fine-tuning on graphs''.
Researchers assume that the structure of graphs can provide hints on what documents are ``semantically similar'' to what other documents.
For example, papers citing each other in an academic graph can be of similar topics.
These methods adopt vanilla language models that take text as input (\textit{e.g.}, BERT \cite{devlin2018bert} and SciBERT \cite{beltagy2019scibert}) as the base model and fine-tune them with structure signals on the graph \cite{cohan2020specter}.
After that, the LLMs will learn node/edge representations that capture the graph homophily from the text perspective. 
This is the simplest way to utilize LLMs on graphs. However, during encoding, the model itself can only consider text. 

Most methods adopt the two-tower encoding and training pipeline, where the representation of each node is obtained separately and the model is optimized as follows:
\begin{gather}
    \mathbf{h}_{v_i} = \text{LLM}_\theta(d_{v_i}), \quad 
    \min_\theta f(\mathbf{h}_{v_i}, \{\mathbf{h}_{v^+_i}\}, \{\mathbf{h}_{v^-_i}\}).
\end{gather}
Here $v^+_i$ represents the positive nodes to $v_i$, $v^-_i$ represents the negative nodes to $v_i$ and $f(\cdot)$ denotes the pairwise training objective.
Different methods have different strategies for $v^+_i$ and $v^-_i$ with different training objectives $f(\cdot)$.
SPECTER \cite{cohan2020specter} constructs the positive text/node pairs with the citation relation, explores random negatives and structure hard negatives, and fine-tunes SciBERT \cite{beltagy2019scibert} with the triplet loss. 
SciNCL \cite{ostendorff2022neighborhood} extends SPECTER by introducing more advanced positive and negative sampling methods based on embeddings trained on graphs.
Touchup-G \cite{zhu2023touchup} proposes the measurement of feature homophily on graphs and brings up a binary cross-entropy fine-tuning objective.
TwHIN-BERT \cite{zhang2023twhin} mines positive node pairs with off-the-shelf heterogeneous information network embeddings and trains the model with a contrastive social loss.
MICoL \cite{zhang2022metadata} discovers semantically positive node pairs with meta-path \cite{sun2011pathsim} and adopts the InfoNCE objective.
E2EG \cite{dinh2022e2eg} utilizes a similar philosophy from GIANT \cite{chien2021node} and adds a neighbor prediction objective apart from the downstream task objective.
WalkLM \cite{tan2023walklm} conducts random walks for structure linearization before fine-tuning the language model.
A summarization of the two-tower graph-centric LLM fine-tuning objectives can be found in Appendix Table \ref{tab:graph-finetune-objective}.

There are other methods using the one-tower pipeline, where node pairs are concatenated and encoded together:
\begin{gather}
    \mathbf{h}_{v_i, v_j} = \text{LLM}_\theta(d_{v_i}, d_{v_j}), \quad
    \min_\theta f(\mathbf{h}_{v_i, v_j}).
\end{gather}
LinkBERT \cite{yasunaga2022linkbert} proposes a document relation prediction objective (an extension of next sentence prediction in BERT \cite{devlin2018bert}) which aims to classify the relation of two node text pairs from contiguous, random, and linked.
MICoL \cite{zhang2022metadata} explores predicting the node pairs' binary meta-path or meta-graph indicated relation with the one-tower language model.

\vspace{-0.05in}
\subsubsection{Discussion}
Although the community is making good progress, there are still some open questions to be solved.

\noindent\textbf{Graph as Code Sequence.} 
Existing graphs as sequence methods are mainly rule-based or GNN-based.
The former relies on natural language to describe the graphs which is not natural for structure data, while the latter has a GNN component that needs to be trained.
A more promising way is to obtain a structure-aware sequence for graphs that can support zero-shot inference.
A potential solution is to adopt codes (that can capture structures, \textit{e.g.}, graph XML or JSON) to describe the graphs and utilize code LLMs \cite{wang2023codet5+}.

\noindent\textbf{Advanced Graph-Empowered LLM Techniques.}
Graph-empowered LLM is a promising direction to achieve foundational models for graphs.
However, existing works are far from enough:
1) \textit{Task}. Existing methods are mainly designed for representation learning (with encoder-only LLMs) which are hard to adopt for generation tasks.
A potential solution is to design Graph-Empowered LLMs with decoder-only or encoder-decoder LLMs as the base architecture.
2) \textit{Pretraining}. Pretraining is important to enable LLMs with contextualized data understanding capability, which can be generalized to other tasks.
However, existing works mainly focus on pretraining LLMs on homogeneous text-attributed networks.
Future studies are needed to explore LLM pretraining in more diverse real-world scenarios including heterogeneous text-attributed networks \cite{jin2023heterformer}, dynamic text-attributed networks \cite{zhang2023llm4dyg}, and textual-edge networks \cite{jin2023edgeformers}.

\vspace{-0.1in}
\subsection{LLM as Encoder}
\begin{figure*}
\centering
\includegraphics[width=16cm]{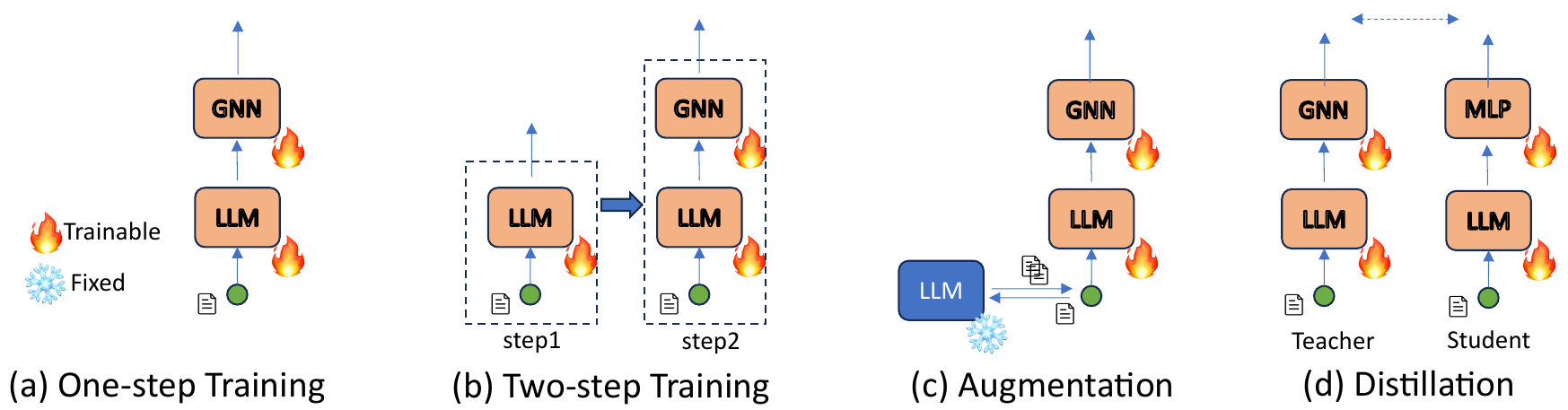}
\vspace{-0.15in}
\caption{The illustration of various techniques related to LLM as Encoder, including (a) One-step Training, (b) Two-step Training, (c) Data Augmentation, and (d) Knowledge Distillation. }\label{fig::gnn-centric}
\vspace{-0.2in}
\end{figure*}
LLMs extract textual features to serve as initial node feature vectors for GNNs, which then generate node/edge representations and make predictions.
These methods typically adopt an LLM-GNN cascaded architecture to obtain the final representation $\mathbf{h}_{v_i}$ for node $v_i$:
\begin{gather}
    \mathbf{x}_{v_i} = \text{LLM}(d_{v_i}) \quad  
    \mathbf{h}_{v_i} = \text{GNN}(\mathbf{X}_v, \mathcal{G}).
\end{gather}
Here $\mathbf{x}_{v_i}$ is the feature vector that captures the textual information $d_{v_i}$ associated with $v_i$.
The final representation $\mathbf{h}_{v_i}$ will contain both textual information and structure information of $v_i$ and can be used for downstream tasks.
In the following sections, we will discuss the optimization, augmentation, and distillation of such models.
The figures for these techniques can be found in Fig. \ref{fig::gnn-centric}.

\vspace{-0.1in}
\subsubsection{Optimization}
\noindent\textbf{One-Step Training} refers to training the LLM and GNN together in the cascaded architecture for the downstream tasks. 
TextGNN \cite{zhu2021textgnn} explores GCN \cite{kipf2016semi}, GraphSAGE \cite{hamilton2017inductive}, GAT \cite{velivckovic2017graph} as the base GNN architecture, adds skip connection between LLM output and GNN output, and optimizes the whole architecture for sponsored search task.
AdsGNN \cite{li2021adsgnn} further extends TextGNN by proposing edge-level information aggregation.
GNN-LM \cite{meng2021gnn} adds GNN layers to enable the vanilla language model to reference similar contexts in the corpus for language modeling.
Joint training LLMs and GNNs in a cascaded pipeline is convenient but may suffer from efficiency \cite{ioannidis2022efficient} (only support sampling a few one-hop neighbors regarding memory complexity) and local minimal \cite{duan2023simteg} (LLM underfits the data) issues.

\vspace{0.3em}
\noindent\textbf{Two-Step Training} means first adapting LLMs to the graph, and then finetuning the whole LLM-GNN cascaded pipeline.
GIANT \cite{chien2021node} proposes to conduct neighborhood prediction with the use of XR-Transformers \cite{zhang2021fast} and results in an LLM that can output better feature vectors than bag-of-words and vanilla BERT \cite{devlin2018bert} embedding for node classification.
LM-GNN \cite{ioannidis2022efficient} introduces graph-aware pre-fine-tuning to warm up the LLM on the given graph before fine-tuning the whole LLM-GNN pipeline and demonstrating significant performance gain.
SimTeG \cite{duan2023simteg} finds that the simple framework of first training the LLMs on the downstream task and then fixing the LLMs and training the GNNs can result in outstanding performance. They further find that using the efficient fine-tuning method, \textit{e.g.}, LoRA \cite{hu2021lora} to tune the LLM can alleviate overfitting issues.
GaLM \cite{xie2023graph} explores ways to pretrain the LLM-GNN cascaded architecture.
The two-step strategy can effectively alleviate the insufficient training of the LLM which contributes to higher text representation quality but is more computationally expensive and time-consuming than the one-step training strategy.

\vspace{-0.1in}
\subsubsection{Data Augmentation}
With its demonstrated zero-shot capability~\cite{wei2021finetuned}, LLMs can be used for data augmentation to generate additional text data for the LLM-GNN cascaded architecture.
The philosophy of using LLM to generate pseudo data is widely explored in NLP~\cite{meng2022generating,jin2023adversarial}.
LLM-GNN \cite{chen2023label} proposes to conduct zero-shot node classification on text-attributed networks by labeling a few nodes and using the pseudo labels to fine-tune GNNs.
TAPE \cite{he2023explanations} presents a method that uses LLM to generate prediction text and explanation text, which serve as augmented text data compared with the original text data. A following medium-scale language model is adopted to encode the texts and output features for augmented texts and original text respectively before feeding into GNNs.
ENG \cite{yu2023empower} brings forward the idea of generating labeled nodes for each category, adding edges between labeled nodes and other nodes, and conducting semi-supervised GNN learning for node classification.

\vspace{-0.1in}
\subsubsection{Knowledge Distillation}
LLM-GNN cascaded pipeline is capable of capturing both text information and structure information. 
However, the pipeline suffers from time complexity issues during inference, since GNNs need to conduct neighbor sampling and LLMs need to encode the text associated with both the center node and its neighbors.
A straightforward solution is to serve the LLM-GNN cascade pipeline as the teacher model and distill it into an LLM as the student model.
In this case, during inference, the model (which is a pure LLM) only needs to encode the text on the center node and avoid time-consuming neighbor sampling.
AdsGNN \cite{li2021adsgnn} proposes an L2-loss to force the outputs of the student model to preserve topology after the teacher model is trained.
GraD \cite{mavromatis2023train} introduces three strategies including the distillation objective and task objective to optimize the teacher model and distill its capability to the student model.

\vspace{-0.1in}
\subsubsection{Discussion}
Given that GNNs are demonstrated as powerful models in encoding graphs, ``LLMs as encoders'' seems to be the most straightforward way to utilize LLMs on graphs.
However, there are still open questions.

\noindent\textbf{Limited Task: Go Beyond Representation Learning.} Current ``LLMs as encoders'' methods or LLM-GNN cascaded architectures are mainly focusing on representation learning, given the single embedding propagation-aggregation mechanism of GNNs, which prevents it from being adopted to generation tasks (\textit{e.g.}, node/text generation).
A potential solution to this challenge can be to conduct GNN encoding for LLM-generated token-level representations and to design proper decoders that can perform generation based on the LLM-GNN cascaded model outputs.

\noindent\textbf{Low Efficiency: Advanced Knowledge Distillation.} The LLM-GNN cascaded pipeline suffers from time complexity issues since the model needs to conduct neighbor sampling and then embedding encoding for each neighboring node.
Although there are methods that explore distilling the learned LLM-GNN model into an LLM model for fast inference, they are far from enough given that the inference of LLM itself is time-consuming.
A potential solution is to distill the model into a much smaller LM or even an MLP. Similar methods \cite{zhang2021graph} have been proven effective in GNN to MLP distillation and are worth exploring for the LLM-GNN cascaded pipeline as well.

\vspace{-0.15in}
\subsection{LLM as Aligner}
\begin{figure}
\centering
\includegraphics[width=8.5cm]{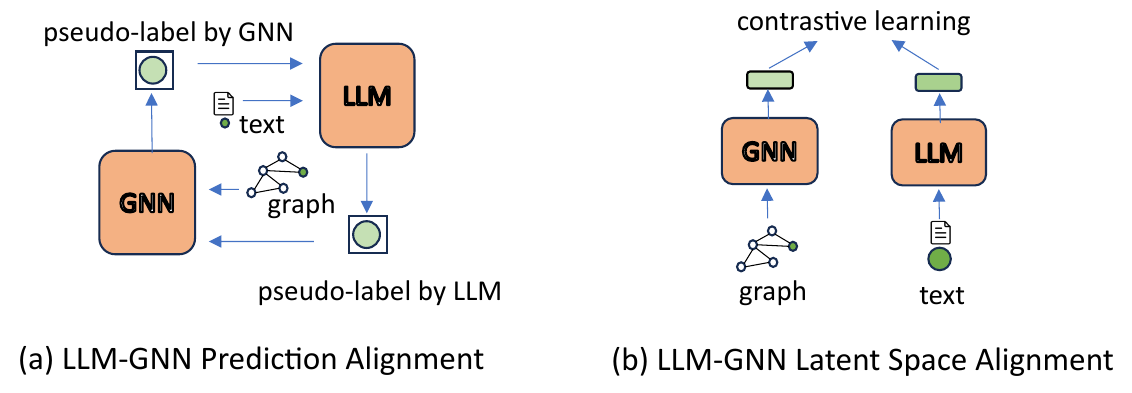}
\vspace{-0.15in}
\caption{The illustration of LLM as Aligner methods, including (a) LLM-GNN Prediction Alignment and (b) LLM-GNN Latent Space Alignment.}\label{fig::gnn-lm}
\vspace{-0.2in}
\end{figure}

These methods contain an LLM component for text encoding and a GNN component for structure encoding. These two components are served equally and trained iteratively or parallelly.
LLMs and GNNs can mutually enhance each other since the LLMs can provide textual signals to GNNs, while the GNNs can deliver structure information to LLMs.
According to how the LLM and the GNN interact, these methods can be further categorized into: LLM-GNN Prediction Alignment and LLM-GNN Latent Space Alignment.
The illustration of these two categories of methods can be found in Fig. \ref{fig::gnn-lm}.

\vspace{-0.1in}
\subsubsection{LLM-GNN Prediction Alignment} 
This refers to training the LLM with the text data on a graph and training the GNN with the structure data on a graph iteratively.
LLM will generate labels for nodes from the text perspective and serve them as pseudo-labels for GNN training, while GNN will generate labels for nodes from the structure perspective and serve them as pseudo-labels for LLM training.
By this design, these two modality encoders can learn from each other and contribute to a final joint text and graph encoding.
In this direction, LTRN \cite{zhang2021minimally} proposes a novel GNN architecture with personalized PageRank \cite{haveliwala2002topic} and attention mechanism for structure encoding while adopting BERT \cite{devlin2018bert} as the language model. The pseudo labels generated by LLM and GNN are merged for the next iteration of training. 
GLEM \cite{zhao2022learning} formulates the iterative training process into a pseudo-likelihood variational framework, where the E-step is to optimize LLM and the M-step is to train the GNN.

\vspace{-0.1in}
\subsubsection{LLM-GNN Latent Space Alignment} 
It denotes connecting text encoding (LLM) and structure encoding (GNN) with cross-modality contrastive learning:
\begin{gather}
\scriptsize
    \mathbf{h}_{d_{v_i}} = \text{LLM}(d_{v_i}),
    \mathbf{h}_{v_i} = \text{GNN}(\mathcal{G}_v), \\
    l(\mathbf{h}_{d_{v_i}}, \mathbf{h}_{{v_i}}) = \frac{\text{Sim}(\mathbf{h}_{d_{v_i}}, \mathbf{h}_{{v_i}})}{\sum_{j\neq i}\text{Sim}(\mathbf{h}_{d_{v_i}}, \mathbf{h}_{{v_j}})}, \\
    \mathcal{L} = \sum_{v_i\in \mathcal{G}} \frac{1}{2|\mathcal{G}|}(l(\mathbf{h}_{d_{v_i}}, \mathbf{h}_{{v_i}}) + l(\mathbf{h}_{{v_i}}, \mathbf{h}_{d_{v_i}}))
\end{gather}
A similar philosophy is widely used in vision-language joint modality learning \cite{radford2021learning}.
Along this line of approaches, ConGrat \cite{brannon2023congrat} adopts GAT \cite{velivckovic2017graph} as the graph encoder and tries MPNet \cite{song2020mpnet} as the language model encoder. 
They have expanded the original InfoNCE loss by incorporating graph-specific elements. These elements pertain to the most likely second, third, and subsequent choices regarding the nodes from which a text originates and the texts that a node generates.
In addition to the node-level multi-modality contrastive objective, GRENADE \cite{li2023grenade} proposes KL-divergence-based neighbor-level knowledge alignment: minimize the neighborhood similarity distribution calculated between LLM and GNN.
G2P2 \cite{wen2023augmenting} further extends node-text contrastive learning by adding text-summary interaction and node-summary interaction. Then, they introduce using label texts in the text modality for zero-shot classification, and using soft prompts for few-show classification.
THLM \cite{zou2023pretraining} proposes to pretrain the language model by contrastive learning with a heterogeneous GNN on heterogeneous text-attributed networks. The pretrained LLM can be fine-tuned on downstream tasks.

\vspace{-0.1in}
\subsubsection{Discussion}

Most existing methods adopt homogeneous text-graph alignment, assuming that the semantic relation between the two modalities, namely text and graph, is singular.
However, this is not usually the case in the real world, given:
1) The existence of multimodal attributes: Other modalities, \textit{e.g.}, images can appear together with text and graph. In this case, it is worth researching how to align the multimodal attributes in a graph scenario.
2) Heterogeneous semantic relations: the semantic relationships between data units (text/image/graph) can be multiplex. Different relations have different distributions and a single semantic alignment will fail to capture the comprehensively \cite{jin2023learning}.

\vspace{-0.1in}
\section{Text-Paired Graphs}\label{sec:graph-text}

Graphs are prevalent data objects in scientific disciplines such as cheminformatics~\cite{pubchem2019,graphseq,wang2021cramolr}, 
material informatics~\cite{pi1m}, bioinformatics~\cite{lai2023keblm}, 
and computer vision~\cite{gcv}.
Within these diverse fields, graphs frequently come paired with critical graph-level text information. 
For instance, molecular graphs in cheminformatics are annotated with text properties such as toxicity, water solubility, and permeability properties~\cite{pubchem2019,pi1m}.
Research on such graphs (scientific discovery) could be accelerated by the text information and the adoption of LLMs. 
In this section, we review the application of LLMs on graph-captioned graphs with a focus on molecular graphs. 
According to the technique categorization in Section \ref{sec:categorizatoin-technique}, we begin by investigating methods that utilize LLMs as Predictor.
Then, we discuss methods that align GNNs with LLMs.
We summarize all surveyed methods in Appendix Table~\ref{tab:graph-text-model-v2} and Figure~\ref{fig::text-graph-pair-methods}.

\vspace{-0.1in}
\subsection{LLM as Predictor}\label{subsec:graph-text-LM-Centric}
In this subsection, we review how to conduct ``LLM as Predictor'' for graph-level tasks. 
Existing methods can be categorized into Graph as Sequence (treat graph data as sequence input) and Graph-Empowered LLMs (design model architecture to encode graphs).

\vspace{-0.1in}
\subsubsection{Graph as Sequence}\label{6.graph-as-sequence}
For text-paired graphs, we have three steps to utilize existing LLM for graph inputs.
\textbf{Step 1}: Linearize graphs into sequence with rule-based methods.
\textbf{Step 2}: Tokenize the linearized sequence.
\textbf{Step 3}: Train/Finetune different LLMs (\textit{e.g.}, Encoder-only, Encoder-Decoder, Decoder-only) for specific tasks. 
We will discuss each step as follows.

\vspace{0.3em}
\noindent\textbf{Step 1: Rule-based Graph Linearization.}
Rule-based linearization converts molecular graphs into text sequences that can be processed by LLMs. To achieve this, researchers develop specifications based on human expertise in the form of line notations~\cite{smiles}. 
For example, the Simplified Molecular-Input Line-Entry System (SMILES)~\cite{smiles} records the symbols of nodes encountered during a depth-first traversal of a molecular graph. 
The International Chemical Identifier (InChI)~\cite{inchi} encodes molecular structures into unique string texts with more hierarchical information. 
Canonicalization algorithms produce unique SMILES for each molecule, often referred to as canonical SMILES. 
However, there are more than one SMILES corresponding to a single molecule and SMILES sometimes represent invalid molecules; LLMs learned from these linearized sequences can easily generate invalid molecules (\textit{e.g.}, incorrect ring closure symbols and unmatched parentheses) due to syntactical errors. 
To this end, DeepSMILES~\cite{deepsmiles} is proposed. It can alleviate this issue in most cases but does not guarantee 100\% robustness. 
The linearized string could still violate basic physical constraints. 
To fully address this problem, SELFIES~\cite{selfies} is introduced which consistently yields valid molecular graphs.

\vspace{0.3em}
\noindent\textbf{Step 2: Tokenization.}
These approaches for linearized sequences are typically language-independent. 
They operate at both character level~\cite{galatica,molca} and substring level~\cite{mfbert,kvplm, momu, momuv2, molxpt, molfm}, based on SentencePiece or BPE~\cite{sentencepiece}. 
Additionally, RT~\cite{regt} proposes a tokenization approach that facilitates handling regression tasks within LM Transformers.

\vspace{0.3em}
\noindent\textbf{Step 3: Encoding the Linearized Graph with LLMs}.
\noindent\textit{Encoder-only LLMs}. 
Earlier LLMs like SciBERT~\cite{beltagy2019scibert} and BioBERT~\cite{biobert} are trained on scientific literature to understand natural language descriptions related to molecules but are not capable of comprehending molecular graph structures. 
To this end, SMILES-BERT~\cite{smilesbert} and MFBERT~\cite{mfbert} are proposed for molecular graph classification with linearized SMILES strings.
Since scientific natural language descriptions contain human expertise which can serve as a supplement for molecular graph structures, recent advances emphasize joint understanding of them~\cite{kvplm, catberta}: 
The linearized graph sequence is concatenated with the raw natural language data and then input into the LLMs. 
Specifically, KV-PLM~\cite{kvplm} is built based on BERT~\cite{devlin2018bert} to understand the molecular structure in a biomedical context. CatBERTa~\cite{catberta}, as developed from RoBERTa~\cite{liu2019roberta}, specializes in the prediction of catalyst properties for molecular graphs.

\vspace{0.1em}
\noindent\textit{Encoder-Decoder LLMs}.
Encoder-only LLMs may lack the capability for generation tasks. 
In this section, we discuss LLMs with encoder-decoder architectures.
For example, Chemformer~\cite{chemformer} uses a similar architecture as BART~\cite{lewis2019bart}. 
The representation from the encoder can be used for property prediction tasks, and the whole encoder-decoder architecture can be optimized for molecule generation. 
Others focus on molecule captioning (which involves generating textual descriptions from a molecule) and text-based molecular generation (where a molecular graph structure is generated from a natural description). 
Specifically, MolT5~\cite{edwards2022translation} is developed based on the T5~\cite{raffel2020exploring}, suitable for these two tasks.  
It formulates molecule-text translation as a multilingual problem and initializes the model using the T5 checkpoint. 
The model was pre-trained on two monolingual corpora: the Colossal Clean Crawled Corpus (C4)~\cite{raffel2020exploring} for the natural language modality and one million SMILES~\cite{chemformer} for the molecule modality. 
Text+Chem T5~\cite{tct} extends the input and output domains to include both SMILES and texts, unlocking LLMs for more generation functions such as text or reaction generation. 
ChatMol~\cite{chatmol} exploits the interactive capabilities of LLMs and proposes designing molecule structures through multi-turn dialogs with T5.

\vspace{0.1em}
\noindent\textit{Decoder-only LLMs}.
Decoder-only architectures have been adopted for recent LLMs due to their advanced generation ability.
MolGPT~\cite{molgpt} and MolXPT~\cite{molxpt} are GPT-style models used for molecule classification and generation. 
Specifically, MolGPT~\cite{molgpt} focuses on conditional molecule generation tasks using scaffolds, while MolXPT~\cite{molxpt} formulates the classification task as a question-answering problem with yes or no responses. 
RT~\cite{regt} adopts XLNet~\cite{yang2019xlnet} and focuses on molecular regression tasks. 
It frames the regression as a conditional sequence modeling problem. 
Galactica~\cite{galatica} is a set of LLMs with a maximum of 120 billion parameters, which is pretrained on two million compounds from PubChem~\cite{pubchem2019}. 
Therefore, Galactica could understand molecular graph structures through SMILES. 
With instruction tuning data and domain knowledge, researchers also adapt general-domain LLMs such as LLaMA to recognize molecular graph structures and solve molecule tasks~\cite{llamamol}.
Recent studies also explore the in-context learning capabilities of LLMs on graphs.
LLM-ICL~\cite{llmicl} assesses the performance of LLMs across eight tasks in the molecular domain, ranging from property classification to molecule-text translation. 
MolReGPT~\cite{molregpt} proposes a method to retrieve molecules with similar structures and descriptions to improve in-context learning.
LLM4Mol~\cite{llm4mol} utilizes the summarization capability of LLMs as a feature extractor and combines it with a smaller, tunable LLM for specific prediction tasks. 

\vspace{-0.1in}
\subsubsection{Graph-Empowered LLMs}\label{sec:graph-text-aligned-transformer}
Different from the methods that adopt the original LLM architecture (\textit{i.e.}, Transformers) and input the graphs as sequences to LLMs, graph-empowered LLMs attempt to design LLM architectures that can conduct joint encoding of text and graph structures.
Some works modify the positional encoding of Transformers.
For instance, GIMLET~\cite{zhao2023gimlet} treats nodes in a graph as tokens. It uses one Transformer to manage both the graph structure and text sequence $[v_1, v_2, \dots, v_{|\mathcal{V}|}, s_{|\mathcal{V}|+1}, \dots, s_{|\mathcal{V}|+{|d_{\mathcal{G}}}|}]$, where $v \in \mathcal{V}$ is a node and $s \in d_{\mathcal{G}}$ is a token in the text associated with $\mathcal{G}$. This sequence cannot reflect graph structure. Therefore, a new position encoding (PE) is used to jointly encode graph structures and text sequences. It defines the relative distance between tokens $i$ and $j$ as follows:
\begin{equation}
{ \scriptsize
\operatorname{PE}(i, j)= \begin{cases}i-j & \text {if $i,j \in d_{\mathcal{G}}$,}\\ \operatorname{GSD} (i, j) + \operatorname{Mean}_{e_k \in \operatorname{SP}(i,j)} \mathbf{x}_{e_{k}} & \text {if $i,j \in \mathcal{V}$,} \\ - \infty & \text {if $i \in \mathcal{V}, j \in d_{\mathcal{G}}$,} \\ 0 & \text {if $i \in d_{\mathcal{G}}, j \in \mathcal{V}$.}\end{cases}
}
\end{equation}
$\operatorname{GSD}$ is the graph shortest distance between two nodes, and $\operatorname{Mean}_{k \in \operatorname{SP}(i,j)}$ represents the mean pooling of the edge features $\mathbf{x}_{e_{k}}$ along the shortest path $\operatorname{SP}(i,j)$ between nodes $i$ and $j$. GIMLET~\cite{zhao2023gimlet} adapts bi-directional attention for node tokens and enables texts to selectively attend to nodes. These designs render the Transformer's submodule, which handles the graph part, equivalent to a Graph Transformer~\cite{graphormer}.

Cross-attention is also used to interact representations between graphs and texts. Given the graph hidden state $\mathbf{h}_\mathcal{G}$, its node-level hidden state $\mathbf{H}_v$ and text hidden state $\mathbf{H}_{d_{\mathcal{G}}}$, Text2Mol~\cite{edwards2021text2mol} implemented interaction between representations in the hidden layers of encoders, while Prot2Text~\cite{prot2text} implemented this interaction within the layers of between encoder and decoder $    \mathbf{H}_{d_{\mathcal{G}}} = \text{softmax}\left(\frac{\mathbf{W}_Q \mathbf{H}_{d_{\mathcal{G}}} \cdot (\mathbf{W}_K \mathbf{H}_v)^T}{\sqrt{d_k}}\right) \cdot \mathbf{W}_V\mathbf{H}_v$,
where $\mathbf{W}_Q, \mathbf{W}_K, \mathbf{W}_V$ are trainable parameters that transform the query modality (e.g., sequences) and the key/value modality (e.g., graphs) into the attention space. 
Furthermore, Prot2Text~\cite{prot2text} utilizes two trainable parameter matrices $\mathbf{W}_1$ and $\mathbf{W}_2$ to integrate the graph representation into the sequence representation $\mathbf{H}_{d_{\mathcal{G}}} =\left(\mathbf{H}_{d_{\mathcal{G}}} +\mathbf{1}_{|{d_{\mathcal{G}}}|} \mathbf{h}_\mathcal{G} \mathbf{W}_1\right) \mathbf{W}_2$.

\subsubsection{Discussion}\label{subsubsec:graph-text-LM-discussion}
\noindent\textbf{LLM Inputs with Sequence Prior.}
\emph{The first challenge is that the progress in advanced linearization methods has not progressed in tandem with the development of LLMs}. Emerging around 2020, linearization methods for molecular graphs like SELFIES offer significant grammatical advantages, yet advanced LMs and LLMs from graph machine learning and language model communities might not fully utilize these, as these encoded results are not part of pretraining corpora prior to their proposal. Consequently, recent studies~\cite{llmicl} indicate that LLMs, such as GPT-3.5/4, may be less adept at using SELFIES compared to SMILES. Therefore, the performance of LM-only and LLM-only methods may be limited by the expressiveness of older linearization methods, as there is no way to optimize these hard-coded rules during the learning pipeline of LLMs.
\emph{However, the second challenge remains as the inductive bias of graphs may be broken by linearization}.
Rule-based linearization methods introduce inductive biases for sequence modeling, thereby breaking the permutation invariance assumption inherent in molecular graphs. It may reduce task difficulty by introducing sequence order to reduce the search space. However, it does not mean model generalization. Specifically, there could be multiple string-based representations for a single graph from single or different approaches. Numerous studies~\cite{enumerateSMILES,randomSMILES,augtSMILES} have shown that training on different string-based views of the same molecule can improve the sequential model's performance, as these data augmentation approaches manage to retain the permutation-invariance nature of graphs. These advantages are also achievable with a permutation-invariant GNN, potentially simplifying the model by reducing the need for complex, string-based data augmentation design.

\noindent\textbf{LLM Inputs with Graph Prior.}
Rule-based linearization may be considered less expressive and generalizable compared to the direct graph representation with rich node features, edge features, and the adjacency matrix~\cite{surveystring}. Various atomic features include atomic number, chirality, degree, formal charge, number of hydrogen atoms, number of radical electrons, hybridization state, aromaticity, and presence in a ring. Bond features encompass the bond's type (e.g., single, double, or triple), the bond's stereochemistry (e.g., E/Z or cis/trans), and whether the bond is conjugated~\cite{ogb}. Each feature provides specific information about atomic properties and structure, crucial for molecular modeling and cheminformatics. 
One may directly \textbf{vectorize} the molecular graph structure into binary vectors~\cite{fingerprint} and then apply parameterized Multilayer Perceptrons (MLPs) on the top of these vectors to get the graph representation. These vectorization approaches are based on human-defined rules and vary, such as MACCS, ECFP, and CDK fingerprints\cite{fingerprint}. These rules take inputs of a molecule and output a vector consisting of 0/1 bits. Each bit denotes a specific type of substructure related to functional groups that could be used for various property predictions. Fingerprints consider atoms and structures, but they cannot automatically learn from the graph structure. GNNs could serve as automatic feature extractors to replace or enhance fingerprints. Some specific methods are explored in Section~\ref{sec:graph-text-aligned-transformer}, while the other graph prior such as the eigenvectors of a graph Laplacian and the random walk prior could also be used~\cite{gps}.

\noindent\textbf{LLM Outputs for Prediction.}
LMs like KV-PLM~\cite{kvplm}, SMILES-BERT~\cite{smilesbert}, MFBERT~\cite{mfbert}, and Chemformer~\cite{chemformer} use a prediction head on the output vector of the last layer. These models are finetuned with standard classification and regression losses but may not fully utilize all the parameters and advantages of the complete architecture. In contrast, models like RT~\cite{regt}, MolXPT~\cite{molxpt}, and Text+Chem T5~\cite{tct} frame prediction as a text generation task. These models are trained with either masked language modeling or autoregressive targets, which requires a meticulous design of the context words in the text~\cite{regt}. Specifically, domain knowledge instructions may be necessary to activate the in-context learning ability of LLMs, thereby making them domain experts~\cite{llmicl}. For example, a possible template could be divided into four parts: \{General Description\}\{Task-Specific Description\}\{Question-Answer Examples\}\{Test Question\}.

\noindent\textbf{LLM Outputs for Reasoning.}
Since string representations of molecular graphs usually carry new and in-depth domain knowledge, which is beyond the knowledge of LLMs, recent work~\cite{molregpt,relm,chemcrow} also attempts to utilize the reasoning ability of LLMs, instead of using them as a knowledge source for predicting the property of molecular graphs. ReLM~\cite{relm} utilizes GNNs to suggest top-k candidates, which were then used to construct multiple-choice answers for in-context learning. ChemCrow~\cite{chemcrow} designs the LLMs as the chemical agent to implement various chemical tools. It avoided direct inference in an expertise-intensive domain.

\vspace{-0.15in}
\subsection{LLM as Aligner}\label{subsec:graph-text-LM-GM}
\subsubsection{Latent Space Alignment}

One may directly align the latent spaces of the GNN and LLM through contrastive learning and predictive regularization. Typically, a graph representation from a GNN can be read out by summarizing all node-level representations, and a sequence representation can be obtained from the [CLS] token. 
We first use two projection heads, which are usually MLPs, to map the separate representation vectors from the GNN and LLM into a unified space as $\mathbf{h}_\mathcal{G}$ and $\mathbf{h}_{d_{\mathcal{G}}}$, and then align them within this space.
Specifically,
MoMu~\cite{momu} and MoMu-v2~\cite{momuv2} retrieve two sentences from the corpus for each molecular graph. During training, graph data augmentation was applied to molecular graphs, creating two augmented views. Consequently, there are four pairs of $\mathcal{G}$ and ${d_{\mathcal{G}}}$. For each pair, the contrastive loss for space alignment is as $\ell_{\text{MoMu}}  =-\log \frac{\exp \left(\operatorname{cos}\left(\mathbf{h}_\mathcal{G}, \mathbf{h}_{d_{\mathcal{G}}}\right) / \tau\right)}{\sum_{\tilde{{d_{\mathcal{G}}}} \neq {d_{\mathcal{G}}}} \exp \left(\operatorname{cos}\left(\mathbf{h}_\mathcal{G}, \mathbf{h}_{\tilde{{d_{\mathcal{G}}}}} \right) / \tau\right)}$
where $\tau$ is the temperature hyper-parameter and $\tilde{{d_{\mathcal{G}}}}$ denotes the sequence not paired to the graph $\mathcal{G}$. 
MoleculeSTM~\cite{moleculestm} also applies contrastive learning to minimize the representation distance between a molecular graph $\mathcal{G}$ and its corresponding texts ${d_{\mathcal{G}}}$, while maximizing the distance between the molecule and unrelated descriptions. 
MoleculeSTM~\cite{moleculestm} randomly samples negative graphs or texts to construct negative pairs of $(\mathcal{G}, \tilde{{d}})$ and $(\tilde{\mathcal{G}}, d)$. 
Similarly, MolFM~\cite{molfm} and GIT-Mol~\cite{gitmol} implement contrastive loss with mutual information and negative sampling.
These two methods also use cross-entropy to regularize the unified space with the assumption that randomly permuted graph and text inputs are predictable if they originate from the same molecule. 

However, the aforementioned methods cannot leverage task labels. 
Given a classification label $y$, CLAMP~\cite{clamp} learns to map active molecules ($y=1$) so that they align with the corresponding assay description for each molecular graph $\mathcal{G}$: $\ell_{\text{CLAMP}} = y \log \left( \sigma \left( \tau^{-1} \mathbf{h}_\mathcal{G}^T \mathbf{h}_{{d_{\mathcal{G}}}} \right) \right) + \left(1-y\right) \log \left(1- \sigma \left( \tau^{-1} \mathbf{h}_\mathcal{G}^T \mathbf{h}_{{d_{\mathcal{G}}}} \right)\right) $.
CLAMP~\cite{clamp} requires labels to encourage that active molecules and their corresponding text descriptions are clustered together in the latent space. 
To advance the alignment between two modalities, MolCA~\cite{molca} trains the Query Transformer (Q-Former)~\cite{qformer} for molecule-text projecting and contrastive alignment. 
Q-former initializes $N_q$ learnable query tokens $\{\mathbf{q}_k\}_{k=1}^{N_q}$. 
These query tokens are updated with self-attention and interact with the output of GNNs through cross-attention to obtain the $k$-th queried molecular representation vector $(\mathbf{h}_\mathcal{G})_k:= \operatorname{Q-Former}(\mathbf{q}_k)$. 
The query tokens share the same self-attention modules with the texts, but use different MLPs, allowing the Q-Former to be used for obtaining the representation of text sequence $\mathbf{h}_{d_{\mathcal{G}}} := \operatorname{Q-Former}(\mathcal{\text{[CLS]}})$. Then we have $\ell_{\mathrm{MolCA}}=-\ell_{\mathrm{g2t}}- \ell_{\mathrm{t2g}}$, where $\ell_{\mathrm{g2t}}= \log \frac{\exp \left(\max _k \cos \left((\mathbf{h}_\mathcal{G})_k, \mathbf{h}_{d_{\mathcal{G}}}\right) / \tau\right)}{\sum_{\tilde{{d_{\mathcal{G}}}} \neq d_\mathcal{G}} \exp \left(\max _k \cos \left((\mathbf{h}_\mathcal{G})_k, \mathbf{h}_{\tilde{{d_{\mathcal{G}}}}}\right) / \tau\right)}$, and $\ell_{\mathrm{t2g}}= \log \frac{\exp \left(\max _k \cos \left(\mathbf{h}_{d_{\mathcal{G}}}, (\mathbf{h}_\mathcal{G})_k\right) / \tau\right)}{\sum_{\tilde{\mathcal{G}} \neq \mathcal{G}} \exp \left(\max _k \cos \left(\mathbf{h}_{d_{\mathcal{G}}}, (\mathbf{h}_{\tilde{\mathcal{G}}})_k \right) / \tau\right)}$.

\subsubsection{Discussion}
\noindent\textbf{Larger-Scale GNNs.} 
GNNs integrate atomic and graph structural features for molecular representation learning~\cite{grea}. Specifically, Text2Mol~\cite{edwards2021text2mol} utilizes the GCN~\cite{kipf2016semi} as its graph encoder and extracts unique identifiers for node features based on Morgan fingerprints~\cite{fingerprint}. MoMu~\cite{momu}, MoMu-v2~\cite{momuv2}, MolFM~\cite{molfm}, GIT-Mol~\cite{gitmol}, and MolCA~\cite{molca} prefer GIN~\cite{gin} as the backbone, as GIN has been proven to be as expressive and powerful as the Weisfeiler-Lehman graph isomorphism test. As described in Section~\ref{subsec:background}, there has been notable progress in making GNNs deeper, more generalizable, and more powerful since the proposal of the GCN~\cite{kipf2016semi} in 2016 and the GIN~\cite{gin} in 2018. However, most reviewed works~\cite{momu,momuv2,molfm,gitmol,molca} are developed using the GIN~\cite{gin} as a proof of concept for their approaches. These pretrained GINs feature five layers and 300 hidden dimensions. The scale of GNNs may be a bottleneck in learning semantic meaningful representation and there is a risk of over-reliance on one modality, neglecting the other. Therefore, for future large-scale GNN designs comparable to LLMs, scaling up the dimension size and adding deeper layers, 
may be considered. Besides, Transformer encoders~\cite{gps} may also improve the expressive power of deep GNNs.

\noindent\textbf{Generation Decoder with GNNs.} GNNs are often not used as decoders for graph generation. The prevalent decoders are mostly text-based, generating linearized graph structures such as SMILES. These methods may be sensitive to the sequence order in the linearized graph. Generative diffusion models~\cite{liu2024inverse} on graphs could be utilized in future work to design generators with GNNs.

\vspace{-0.15in}
\section{Resources and Applications}\label{sec:application}

\subsection{Datasets, Splitting and Evaluation}

\begin{table*}
    \centering
    \caption{\small Data collection in Section~\ref{sec:graph-node-text} for text-attributed graphs. Task: ``NC'', ``UAP'', ``LP'', ``Rec'', ``EC'', ``RG'' denote node classification, user activity prediction, link prediction, recommendation, edge classification, and regression task.}
    \vspace{-0.15in}
    \label{tab:node_text_data}
    \begin{adjustbox}{width=0.92\textwidth}
    \begin{tabular}{llllllll}
    \toprule
       Text. &  Data & Year & Task & \# Nodes & \# Edges & Domain & Source \& Notes \\
    \midrule
    \parbox[t]{2mm}{\multirow{15}{*}{\rotatebox[origin=c]{90}{Node}}}
    & ogb-arxiv & 2020.5 & NC & 169,343 & 1,166,243 & Academic & OGB \cite{ogb}	 \\
    & ogb-products & 2020.5 & NC & 2,449,029 & 61,859,140 & E-commerce & OGB \cite{ogb} \\
    & ogb-papers110M & 2020.5 & NC & 111,059,956 & 1,615,685,872	 & Academic & OGB \cite{ogb} \\
    & ogb-citation2 & 2020.5 & LP & 2,927,963 & 30,561,187		 & Academic & OGB \cite{ogb} \\
    & Cora   & 2000 & NC & 2,708 & 5,429		 & Academic & \cite{cora} \\ 
    & Citeseer & 1998 & NC & 3,312 & 4,732		 & Academic & \cite{citeseer} \\
    & DBLP & 2023.1 & NC, LP & 5,259,858 & 36,630,661		 & Academic & \url{www.aminer.org/citation} \\
    & MAG & 2020 & NC, LP, Rec  RG & $\sim 10$M & $\sim 50$M		 & Academic & multiple domains \cite{wang2020microsoft}\cite{zhang2023effect} \\
    & Goodreads-books & 2018 & NC, LP & $\sim 2$M & $\sim 20$M		 & Books & multiple domains \cite{wan2018item} \\
    & Amazon-items & 2018 & NC, LP, Rec & $\sim 15.5$M & $\sim 100$M & E-commerce & multiple domains \cite{ni2019justifying}  \\
    & SciDocs & 2020 & NC, UAP, LP, Rec & - & -		 & Academic & \cite{cohan2020specter}  \\ 
    & PubMed & 2020 & NC & 19,717 & 44,338		 & Academic & \cite{sen2008collective}  \\
    & Wikidata5M & 2021 & LP & $\sim 4$M & $\sim 20$M		 & Wikipedia & \cite{wang2021kepler}  \\
    & Twitter & 2023 & NC, LP & 176,279 & 2,373,956		 & Social & \cite{brannon2023congrat}  \\
    \midrule
    \parbox[t]{2mm}{\multirow{3}{*}{\rotatebox[origin=c]{90}{Edge}}}
    & Goodreads-reviews & 2018 & EC, LP & $\sim 3$M & $\sim 100$M		 & Books & multiple domains \cite{wan2018item} \\
    & Amazon-reviews & 2018 & EC, LP & $\sim 15.5$M & $\sim 200$M & E-commerce & multiple domains \cite{ni2019justifying} \\
    & Stackoverflow & 2023 & EC, LP & 129,322 & 281,657 & Social		 & \cite{jin2023edgeformers}  \\
    \bottomrule
    \end{tabular}
    \end{adjustbox}
\vspace{-0.1in}
\end{table*}

\begin{table*}
    \centering
    \caption{Data collection in Section \ref{sec:graph-text} for text-captioned graphs. ``PT'', ``FT'', ``Cap.'', ``GC'', ``Retr.', and ``Gen.'' refer to pretraining, finetuning, caption, graph classification, retrieval, and graph generation, respectively. The superscript for the size denotes \# graph-text pairs$^1$, \# graphs$^2$, \# assays$^3$.}
    \vspace{-0.15in}
    \label{tab:graph-text-data}
    \begin{adjustbox}{width=0.8\textwidth}
    \begin{tabular}{lllll}
    \toprule
    Data & Date & Task & Size & Source \& Notes \\
    \midrule
    ChEMBL-2023~\cite{chembl2023} & 2023 & Various & 2.4M$^2$,20.3M$^3$ & Drug-like \\
    PubChem~\cite{pubchem2019} & 2019 & Various & 96M$^2$,237M$^3$ & Biomedical \\ 
    \multirow{1}{*}{PC324K~\cite{molca}} & \multirow{1}{*}{2023} & PT, Cap., & \multirow{1}{*}{324K$^1$} & \multirow{1}{*}{PubChem~\cite{pubchem2019}} \\
    \multirow{1}{*}{MolXPT-PT~\cite{molxpt}} & \multirow{1}{*}{2023} & \multirow{1}{*}{PT} & \multirow{1}{*}{30M$^2$} & \multirow{1}{*} PubChem~\cite{pubchem2019}, PubMed, ChEBI~\cite{chebi} \\
    ChE-bio~\cite{zhao2023gimlet} & 2023 & PT & 365K$^2$ & ChEMBL~\cite{chembl2012} \\
    ChE-phy~\cite{zhao2023gimlet} & 2023 & PT & 365K$^2$  & ChEMBL~\cite{chembl2012} \\
    ChE ZS~\cite{zhao2023gimlet} & 2023 & GC & 91K$^2$ & ChEMBL~\cite{chembl2012} \\
    \multirow{1}{*}{PC223M~\cite{clamp}} & \multirow{1}{*}{2023} & \multirow{1}{*}{PT, Retr.} & \multirow{1}{*}{223M$^1$,2M$^2$,20K$^3$} & PubChem~\cite{pubchem2019} \\
    PCSTM~\cite{moleculestm} & 2022 & PT & 281K$^1$ & PubChem~\cite{pubchem2019} \\
    \multirow{1}{*}{PCdes~\cite{pubchem2019}}& \multirow{1}{*}{2022} & FT, Cap, Retr. & \multirow{1}{*}{15K$^1$} &
    PubChem~\cite{pubchem2019} \\
    \multirow{1}{*}{ChEBI-20~\cite{edwards2021text2mol}} & \multirow{1}{*}{2021} & FT., Retr., Gen., Cap. &  \multirow{1}{*}{33K$^1$} & PubChem~\cite{pubchem2019}, ChEBI~\cite{chebi} \\
    \bottomrule
    \end{tabular}
    \end{adjustbox}
\vspace{-0.15in}
\end{table*}

We summarize the datasets for three scenarios (namely pure graphs, text-attributed graphs, and text-paired graphs) and show them in Table \ref{tab:reasoning-problems}, Table \ref{tab:node_text_data}, and Table \ref{tab:graph-text-data} respectively.

\vspace{-0.1in}
\subsubsection{Pure Graphs}

In Table~\ref{tab:reasoning-problems}, we summarize the pure graph reasoning problems discussed in Section~\ref{sec:graph-reasoning}.
Many problems are shared or revisited in different datasets due to their commonality.
NLGraph~\cite{wang2023can}, LLMtoGraph~\cite{liu2023evaluating} and GUC~\cite{guo2023gpt4graph} study a set of standard graph reasoning problems, including connectivity, shortest path, and graph diameter.
GraphQA~\cite{fatemi2023talk} benchmarks a similar set of problems but additionally describes the graphs in real-world scenarios to study the effect of graph grounding.
LLM4DyG~\cite{zhang2023llm4dyg} focuses on reasoning tasks on temporally evolving graphs.
Accuracy is the most common evaluation metric as they are primarily formulated as graph question-answering tasks.

\vspace{-0.1in}
\subsubsection{Text-Attributed Graphs}
We summarize the famous datasets for evaluating models on text-attributed graphs in Table \ref{tab:node_text_data}.
The datasets are mostly from the academic, e-commerce, book, social media, and Wikipedia domains.
The popular tasks to evaluate models on those datasets include node classification, link prediction, edge classification, regression, and recommendation.
The evaluation metrics for node/edge classification include Accuracy, Macro-F1, and Micro-F1. For link prediction and recommendation evaluation, Mean Reciprocal Rank (MRR), Normalized Discounted Cumulative Gain (NDCG), and Hit Ratio (Hit) usually serve as metrics. While evaluating model performance on regression tasks, people tend to adopt mean absolute errors (MAE) or root mean square error (RMSE).

\vspace{-0.1in}
\subsubsection{Text-Paired Graphs}
Table~\ref{tab:graph-text-data} shows text-paired graph datasets (including text-available and graph-only datasets).
For \textit{Data Splitting}, options include random splitting, source-based splitting, activity cliffs and scaffolds~\cite{keyelement}, and data balancing~\cite{sgir}. 
Graph classification usually adopts AUC~\cite{ogb} as the metrics, while regression uses MAE, RMSE, and R$^2$~\cite{grea}. 
For text generation evaluation, people tend to use the Bilingual Evaluation Understudy (BLEU) score; while for molecule generation evaluation, heuristic evaluation methods (based on factors including validity, novelty, and uniqueness) are adopted.
However, it is worth noted that BLEU score is efficient but less accurate, while heuristic evaluation methods are problematic subject to unintended modes, such as the superfluous addition of carbon atoms in~\cite{molgenfail}.

\vspace{-0.15in}
\subsection{Open-Source Implementations}
\noindent\textbf{HuggingFace.} HF Transformers\footnote{https://huggingface.co/docs/transformers/index} is the most popular Python library for Transformers-based language models. 
Besides, it also provides two additional packages: Datasets\footnote{https://huggingface.co/docs/datasets/index} for easily accessing and sharing datasets and Evaluate\footnote{https://huggingface.co/docs/evaluate/index} for easily evaluating machine learning models and datasets.

\noindent\textbf{Fairseq.} Fairseq\footnote{https://github.com/facebookresearch/fairseq} is another open-source Python library for Transformers-based language models.

\noindent\textbf{PyTorch Geometric.} PyG\footnote{https://pytorch-geometric.readthedocs.io/en/latest/index.html} is an open-source Python library for graph machine learning. 
It packages more than 60 types of GNN, aggregation, and pooling layers.

\noindent\textbf{Deep Graph Library}. DGL\footnote{https://www.dgl.ai/} is another open-source Python library for graph machine learning.

\noindent\textbf{RDKit}. RDKit\footnote{https://www.rdkit.org/docs/} is one of the most popular open-source cheminformatics software programs that facilitates various operations and visualizations for molecular graphs. 
It offers many useful APIs, such as the linearization implementation for molecular graphs, to convert them into easily stored SMILES and to convert these SMILES back into graphs. 

\vspace{-0.1in}
\subsection{Practical Applications}

\subsubsection{Scientific Discovery}

\noindent\textbf{Virtual Screening.} It aims to search a library of unlabeled molecules to identify useful structures for a given task. Machine learning models could automatically screen out trivial candidates to accelerate this process. However, training accurate models is not easy since labeled molecules are limited in size and imbalanced in distribution~\cite{sgir}. There are many efforts to improve GNNs against data sparsity~\cite{grea,sgir,dct}. However, it is difficult for a model to generalize and understand in-depth domain knowledge that it has never been trained on. Texts could be complementary knowledge sources. Discovering task-related content from massive scientific papers and using them as instructions has great potential to design accurate GNNs in virtual screening~\cite{zhao2023gimlet}. 

\noindent\textbf{Molecular Generation.}
Molecular generation and optimization is one fundamental goal for drug and material discovery. Scientific hypotheses of molecules~\cite{scientificdiscovery}, can be represented in the joint space of GNNs and LLMs. Then, one may search in the latent space for a better hypothesis that aligns with the text description (human requirements) and adheres to structural constraints like chemical validity. Chemical space has been found to contain more than $10^{60}$ molecules~\cite{chemicalspace}, which is beyond the capacity of exploration in wet lab experiments. Generating constrained candidates within relevant subspaces is a challenge~\cite{liu2024inverse} and promising, especially when incorporating textual conditions.

\noindent\textbf{Synthesis Planning.} Synthesis designs start from available molecules and involve planning a sequence of steps that can finally produce a desired chemical compound through a series of reactions~\cite{scientificdiscovery}. This procedure includes a sequence of reactant molecules and reaction conditions. Both graphs and texts play important roles in this process. For example, graphs may represent the fundamental structure of molecules, while texts may describe the reaction conditions, additives, and solvents. LLMs can assist in the planning by suggesting possible synthesis paths directly or by serving as agents to operate on existing planning tools~\cite{chemcrow}.

\vspace{-0.1in}
\subsubsection{Computational Social Science}

In computational social science, researchers are interested in modeling the behavior of people/users and discovering new knowledge that can be utilized to forecast the future.
The behaviors of users and interactions between users can be modeled as graphs, where the nodes are associated with rich text information (\textit{e.g.}, user profile, messages, emails).
We will show two example scenarios below.

\noindent\textbf{E-commerce.}
In E-commerce platforms, there are many interactions (\textit{e.g.}, purchase, view) between users and products.
For example, users can view or purchase products. 
In addition, the users, products, and their interactions are associated with rich text information.
For instance, products have titles/descriptions and users can leave a review of products.
In this case, we can construct a graph \cite{lin2021personalized} where nodes are users and products, while edges are their interactions.
Both nodes and edges are associated with text.
It is important to utilize both the text information and the graph structure information (user behavior) to model users and items and solve complex downstream tasks (\textit{e.g.}, item recommendation \cite{he2020lightgcn}, bundle recommendation \cite{chang2020bundle}, and product understanding \cite{xu2019open}).

\noindent\textbf{Social Media.}
In social media platforms, there are many users and they interact with each other through messages, emails, and so on.
In this case, we can build a graph where nodes are users and edges are the interaction between users.
There will be text associated with nodes (\textit{e.g.}, user profile) and edges (\textit{e.g.}, messages).
Interesting research questions will be how to do joint text and graph structure modeling to deeply understand the users for friend recommendation \cite{chen2020friend}, user analysis \cite{wang2016unsupervised}, community detection \cite{shchur2019overlapping}, and personalized response generation \cite{sun2023decoding,sun2023measuring}. 

\vspace{-0.1in}
\subsubsection{Specific Domains}
In many specific domains, text data are interconnected and lie in the format of graphs. The structure information on the graphs can be utilized to better understand the text unit and contribute to advanced problem-solving.

\noindent\textbf{Academic Domain.} In the academic domain, graphs \cite{wang2020microsoft} are constructed with papers as nodes and their relations (\textit{e.g.}, citation, authorship, etc) as edges. The representation learned for papers on such graphs can be utilized for paper recommendation \cite{bai2019scientific}, paper classification \cite{chowdhury2020research}, and author identification \cite{madigan2005author}.

\noindent\textbf{Legal Domain.} In the legal domain, opinions given by the judges always contain references to opinions given for previous cases. In such scenarios, people can construct a graph \cite{whalen2016legal} based on the citation relations between opinions. The representations learned on such a graph with both text and structure information can be utilized for clause classification \cite{friedrich2016situation} and opinion recommendation \cite{guha2023legalbench}.

\noindent\textbf{Education Domain.} In the education domain, we can construct a graph with coursework as nodes and their relations as edges. The model learned on such a graph can be utilized for knowledge tracing \cite{nakagawa2019graph} and student performance prediction \cite{li2020peer}.

\vspace{-0.1in}
\section{Future directions}\label{sec:future-direction}
\noindent\textbf{Better Benchmark Datasets.} 
Most pure graph benchmarks evaluate LLMs' reasoning ability on homogeneous graphs but do not include evaluations on heterogeneous or spatial-temporal graphs.
For text-attributed graphs, as summarized in Table \ref{tab:node_text_data}, most benchmark datasets are from academic domains and e-commerce domains.
However, in the real world, text-attributed graphs are ubiquitous across multiple domains (\textit{e.g.}, legal and health).
More diverse datasets are needed to comprehensively evaluate LLMs on real-world scenarios.
For text-paired graphs, as summarized in Table~\ref{tab:graph-text-data}, there is a lack of comprehensive datasets covering various machine learning tasks in chemistry. 
Although a massive number of scientific papers are available, preprocessing them into a ready-to-use format and pairing them with specific molecular graph data points of interest remains a cumbersome and challenging task. Besides, we could investigate graph-text pairs in 3D space, where each molecule may be associated with atomic coordinates~\cite{ai4sci}.

\noindent\textbf{Broader Task Space with LLMs.} 
More comprehensive studies on the performance of LLMs for graph tasks hold promise for the future. While LLMs as encoder approaches have been explored for text-attributed graphs, their application to text-captioned molecular graphs remains underexplored. Promising directions include using LLMs for data augmentation and knowledge distillation to design domain-specific GNNs for various text-paired graph tasks. Furthermore, although graph generation has been approached in text-paired graphs, it remains an open problem for text-attributed graphs (\textit{i.e.}, how to conduct joint text and graph structure generation)

\noindent\textbf{Efficienct LLMs on Graphs.} 
While LLMs have shown a strong capability to learn on graphs, they suffer from inefficiency in graph linearization and model optimization.
On one hand, as discussed in Section \ref{5.graph-as-sequence} and \ref{6.graph-as-sequence}, many methods rely on transferring graphs into sequences that can be inputted into LLMs.
However, the length of the transferred sequence will increase significantly as the size of the graph increases.
This poses challenges since LLMs always have a maximum sequence input length and a long input sequence will lead to higher time and memory complexity.
On the other hand, optimizing LLMs itself is computationally expensive.
Although some general efficient tuning methods such as LoRA are proposed, there is a lack of discussion on graph-aware LLM efficient tuning methods.

\noindent\textbf{Generalizable and Robust LLMs on Graphs.} 
Another interesting direction is to explore the generalizability and robustness of LLMs on graphs.
Generalizability refers to having the ability to transfer the knowledge learned from one domain graph to another; while robustness denotes having consistent prediction regarding obfuscations and attacks.
Although LLMs have demonstrated their strong generalizability in processing text, they still suffer from robustness and hallucination issues, which are to be solved for graph data modeling as well.

\noindent\textbf{Multi-Modal Foundation Models.} 
One open question is, ``Should we use one foundation model to unify different modalities, and how?''
The modalities can include texts, graphs, and even images.
For instance, molecules can be represented as graphs, described as texts, and photographed as images; products can be treated as nodes in a graph, associated with a title/description, and combined with an image. 
Designing a model that can conduct joint encoding for all modalities will be useful but challenging.
Furthermore, there has always been tension between building a unified foundational model and customizing model architectures for different domains. It is thus intriguing to ask whether a unified architecture will suit different data types, or if tailoring model designs according to domains will be necessary. Correctly answering this question can save economic and intellectual resources from unnecessary attempts and also shed light on a deeper understanding of graph-related tasks.

\noindent\textbf{LLMs as Dynamic Agents on Graphs.}
Although LLMs have shown their advanced capability in generating text, one-pass generation of LLMs suffers from hallucination and misinformation issues due to the lack of accurate parametric knowledge. Simply augmenting retrieved knowledge in context is also bottlenecked by the capacity of the retriever.
In many real-world scenarios, graphs such as academic networks, and Wikipedia are dynamically looked up by humans for knowledge-guided reasoning. Simulating such a role of dynamic agents can help LLMs more accurately retrieve relevant information via multi-hop reasoning, thereby correcting their answers and alleviating hallucinations.

\vspace{-0.15in}
\section{Conclusion}\label{sec:conclusion}
In this paper, we provide a comprehensive review of large language models on graphs.
We first categorize graph scenarios where LMs can be adopted and summarize the large language models on graph techniques.
We then provide a thorough review, analysis, and comparison of methods within each scenario.
Furthermore, we summarize available datasets, open-source codebases, and multiple applications.
Finally, we suggest future directions for large language models on graphs.

\vspace{-0.2in}
\section*{Acknowledgments}
This work was supported in part by US DARPA INCAS Program No. HR0011-21-C0165 and BRIES Program No. HR0011-24-3-0325, National Science Foundation IIS-19-56151, the Molecule Maker Lab Institute: An AI Research Institutes program supported by NSF under Award No. 2019897, and the Institute for Geospatial Understanding through an Integrative Discovery Environment (I-GUIDE) by NSF under Award No. 2118329,
U.S. DARPA ITM Program No. FA8650-23-C-7316 and Agriculture and Food Research Initiative (AFRI) grant no. 2020-67021- 32799/project accession no.1024178 from the USDA National Institute of Food and Agriculture. This work was also supported by NSF under Award No. 2142827, 2146761, 2234058, and by ONR N00014-22-1-2507.
Any opinions, findings, and conclusions or recommendations expressed herein are those of the authors and do not necessarily represent the views, either expressed or implied, of DARPA or the U.S. Government.

\vspace{-50pt}
\begin{IEEEbiography}[{\includegraphics[width=1in,
height=1.25in,clip,keepaspectratio]{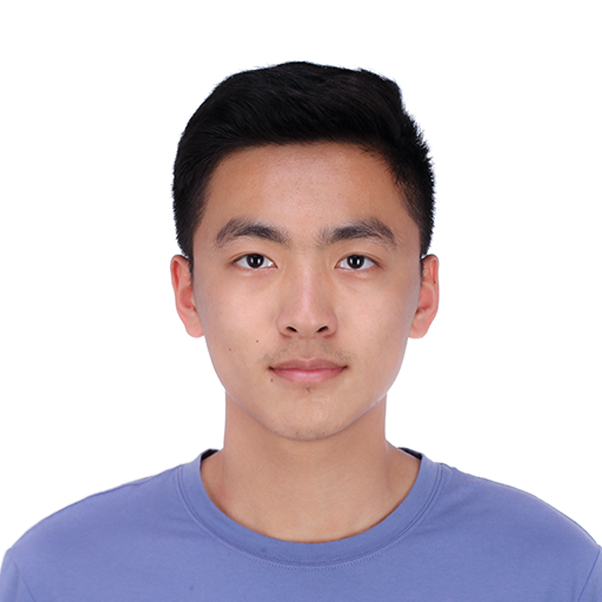}}]{Bowen Jin} is a Ph.D.~candidate in computer science at the University of Illinois at Urbana-Champaign, advised by Prof. Jiawei Han. He received his B.S. degree at Tsinghua University in 2021. 
His research focuses on large language models, information networks, and data/text mining, with their applications in information retrieval and knowledge discovery.
He has published first-authored papers in SIGIR, ICLR, ACL, and KDD.
He receives the Apple PhD Fellowship in 2024.
\end{IEEEbiography}
\vspace{-40pt}
\begin{IEEEbiography}[{\includegraphics[width=1in,height=1.25in,clip,keepaspectratio]{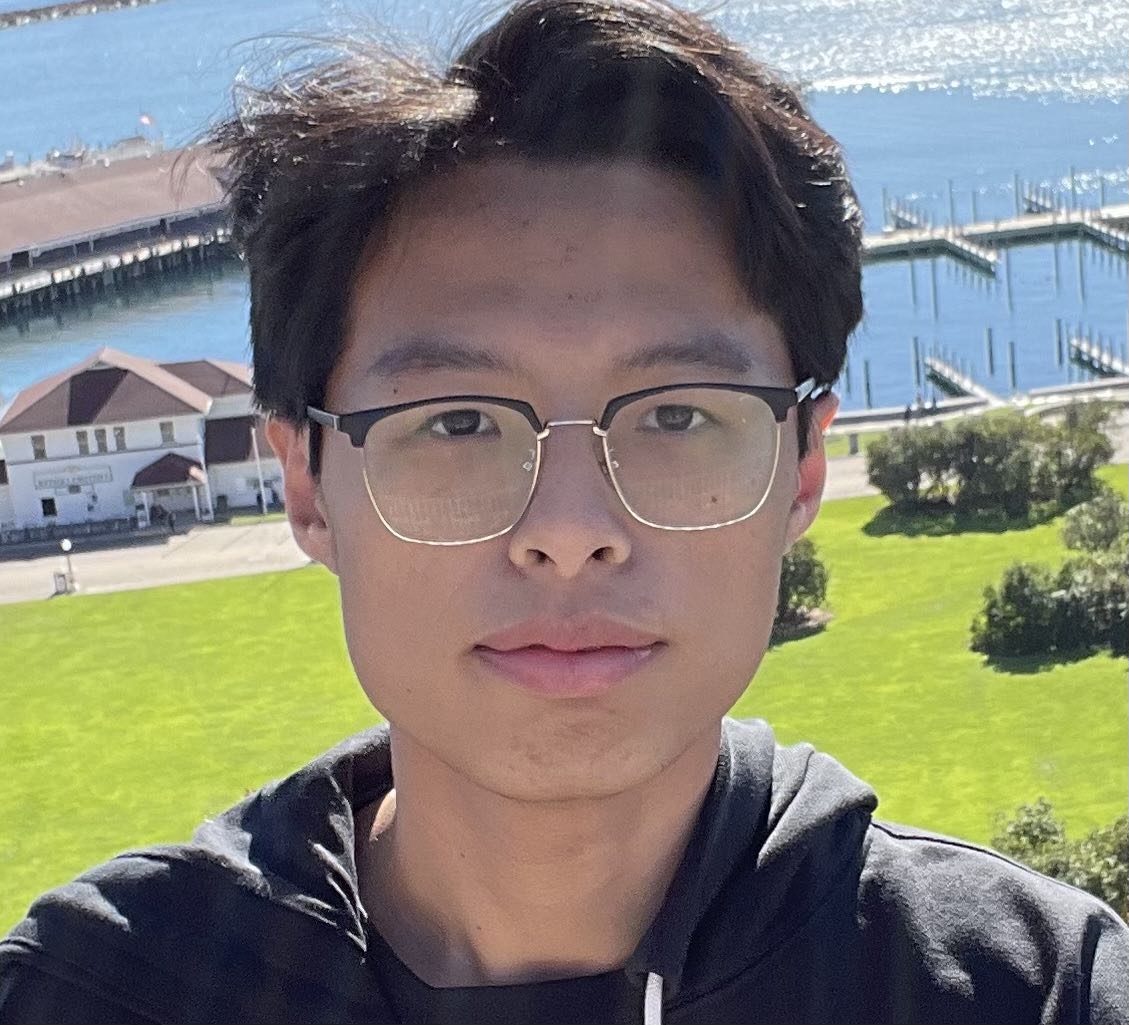}}]{Gang Liu} is a Ph.D.~student of Computer Science and Engineering at the University of Notre Dame, advised by Prof. Meng Jiang. He received his B.S. degree at Southwest University in 2021. His research interest is graph machine learning (e.g., prediction and generation) with applications in scientific discovery (e.g., molecules, polymers). He has first-authored publications in top venues like KDD, NeurIPS, and TKDD.
\end{IEEEbiography}
\vspace{-50pt}
\begin{IEEEbiography}[{\includegraphics[width=1in,height=1.25in,clip,keepaspectratio]{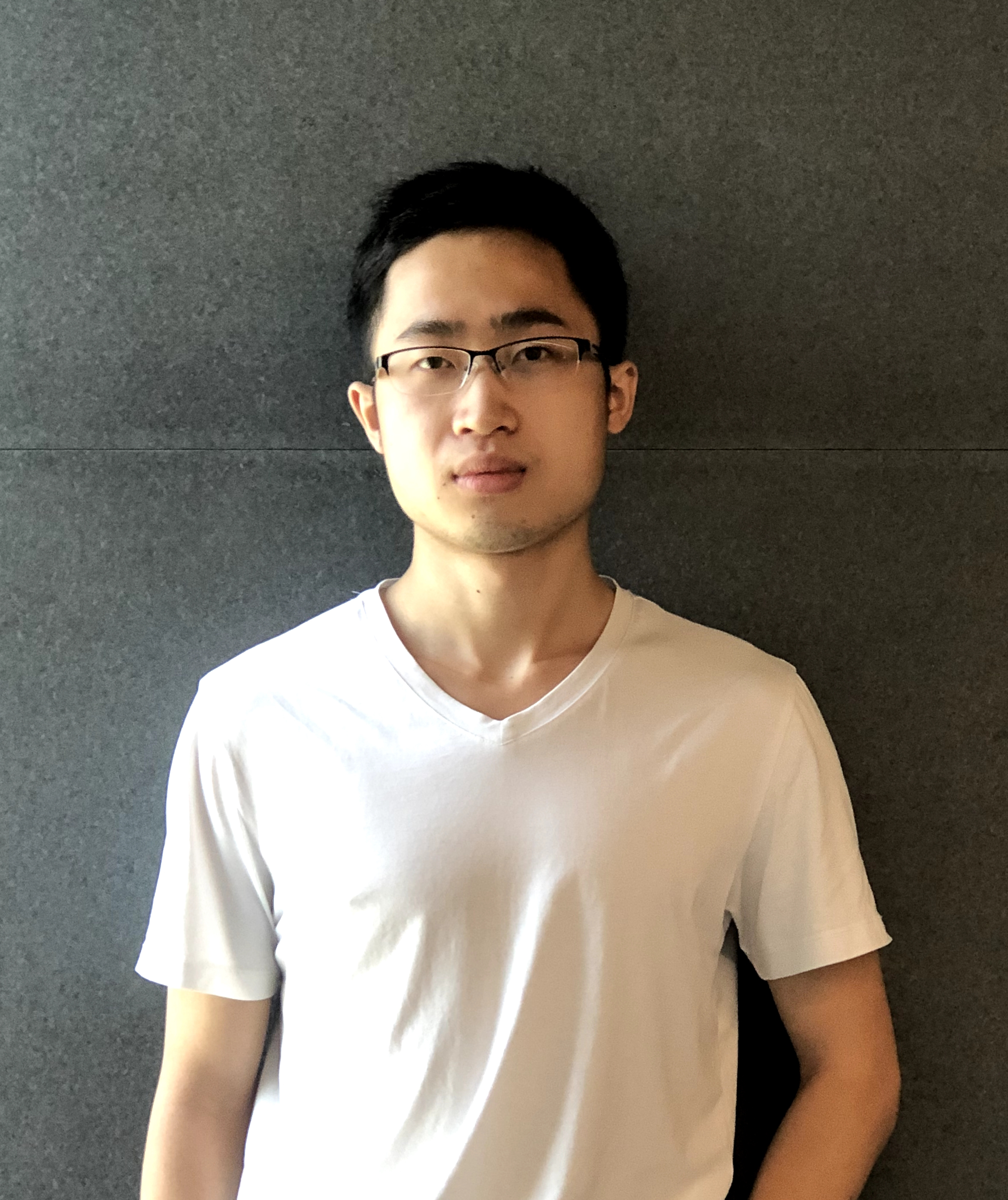}}]{Chi Han} is a Ph.D.~candidate in computer science at the University of Illinois at Urbana-Champaign, advised by Prof. Heng Ji. His research interests are centered around understanding of large language models (LLMs) representations to provide insights into and develop useful adaptations for LLMs. He has published first-author papers in NeurIPS, ICLR NAACL and ACL. His work received Outstanding Paper Awards in NAACL 2024 and ACL 2024.
\end{IEEEbiography}
\vspace{-50pt}
\begin{IEEEbiography}[{\includegraphics[width=1in,height=1.25in,clip,keepaspectratio]{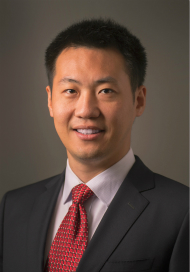}}]{Meng Jiang} received his B.E and Ph.D. from Tsinghua University in 2010 and 2015. He spent two years in UIUC as a postdoc and joined the faculty of University of Notre Dame in 2017, where he is currently an Associate Professor of Computer Science and Engineering. His research interests include data mining, machine learning, and natural language processing. He was given by NSF the CAREER award in 2022. The honors and awards he received include Best Paper Finalist in KDD 2014, Best Paper Award in KDD-DLG workshop 2020, ACM SIGSOFT Distinguished Paper Award in ICSE 2021, and Outstanding Paper Award in EMNLP 2023.
\end{IEEEbiography}
\vspace{-40pt}
\begin{IEEEbiography}
[{\includegraphics[width=1in,height=1.25in,clip,keepaspectratio]{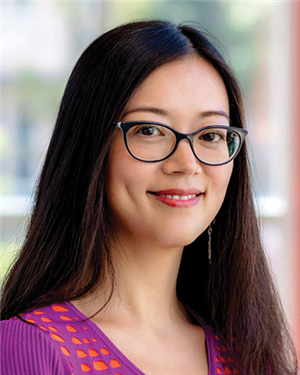}}]{Heng Ji} is a professor at Computer Science Department, and an affiliated faculty member at Electrical and Computer Engineering Department and Coordinated Science Laboratory of University of Illinois Urbana-Champaign. She is an Amazon Scholar and the Founding Director of Amazon-Illinois Center on AI for Interactive Conversational Experiences. Her research interests focus on Natural Language Processing, especially on Multimedia Multilingual Information Extraction, Knowledge-enhanced Large Language Models, Knowledge-driven Generation, and Conversational AI. She was selected as ``Young Scientist'' by the World Economic Forum in 2016 and 2017 and was named as part of Women Leaders of Conversational AI (Class of 2023) by Project Voice. The awards she received include ``AI's 10 to Watch'' Award by IEEE Intelligent Systems in 2013, NSF CAREER award in 2009, PACLIC2012 Best paper runner-up, ``Best of ICDM2013'' paper award, ``Best of SDM2013'' paper award, ACL2020 Best Demo Paper Award, NAACL2021 Best Demo Paper Award, Google Research Award in 2009 and 2014, IBM Watson Faculty Award in 2012 and 2014 and Bosch Research Award in 2014-2018. 
\end{IEEEbiography}
\vspace{-40pt}
\begin{IEEEbiography}[{\includegraphics[width=1in,height=1.25in,clip,keepaspectratio]{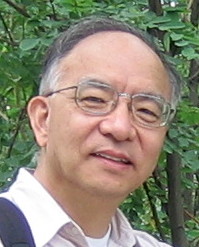}}]{Jiawei Han} is Michael Aiken Chair Professor in the Siebel School of Computing and Data Science, University
of Illinois Urbana-Champaign. He has been researching into data mining, text mining, machine learning, and large language models, with over 1000 publications. He served as the founding Editor-in-Chief of ACM Transactions on Knowledge Discovery from Data (TKDD) (2007-2012). Jiawei has received ACM SIGKDD Innovation Award (2004), IEEE 
Computer Society Technical Achievement Award (2005), IEEE Computer Society W.~Wallace McDowell 
Award (2009), and Japan's Funai Achievement Award (2018). He is Fellow of ACM and Fellow of IEEE and served as co-Director of KnowEnG, a Center of Excellence in Big Data Computing (2014-2019), 
funded by NIH Big Data to Knowledge (BD2K) Initiative and as the Director of Information Network 
Academic Research Center (INARC) (2009-2016) supported by the Network Science-Collaborative 
Technology Alliance (NS-CTA) program of U.S.~Army Research Lab. His co-authored textbook ``Data Mining: Concepts and Techniques'' (Morgan Kaufmann) has been adopted popularly as a textbook worldwide.
\end{IEEEbiography}

\newpage
\appendix

\subsection{Training \& Inference Framework with LLMs}
\label{app:training}
There are two typical training and inference paradigms to apply language models on graphs: 1) Pretraining-then-finetuning: typically adopted for medium-scale large language models; and 2) Pretraining-then-prompting: typically adopted for large-scale large language models.

\noindent\textbf{Pretraining} denotes training the language model with unsupervised objectives to initialize them with language understanding and inference ability for downstream tasks.
Typical pretraining objectives for pure text include masked language modeling \cite{devlin2018bert}, auto-regressive causal language modeling \cite{brown2020language}, corruption-reconstruction language modeling \cite{lewis2019bart} and text-to-text transfer modeling \cite{raffel2020exploring}. 
When extended in the graph domain, language model pretraining strategies include document relation prediction \cite{yasunaga2022linkbert}, network-contextualized masked language modeling \cite{jin2023patton}, contrastive social prediction \cite{zhang2022twhin} and context graph prediction \cite{zou2023pretraining}.

\noindent\textbf{Finetuning} refers to the process of training the language model with labeled data for the downstream tasks. Language model fine-tuning methodology can be further categorized into fully fine-tuning, efficient fine-tuning, and instruction tuning.

\begin{itemize}
    \item \textbf{Full Finetuning} means updating all the parameters inside the language model. It is the most commonly used fine-tuning method that fully stimulates the language model's potential for downstream tasks, but can suffer from heavy computational overload \cite{kasneci2023chatgpt} and result in overfitting issues\cite{duan2023simteg}.
    \item \textbf{Efficient Finetuning} refers to only fine-tuning a subset of parameters inside the language model. Efficient tuning methods for pure text include prompt tuning \cite{lester2021power}, prefix tuning \cite{li2021prefix}, adapter \cite{houlsby2019parameter} and LoRA \cite{hu2021lora}. Efficient language model fine-tuning methods particularly designed for graph data include graph neural prompt \cite{tian2023graph} and graph-enhanced prefix \cite{chai2023graphllm}.
    \item \textbf{Instruction Tuning} denotes fine-tuning language model with downstream task instructions \cite{wei2021finetuned}\cite{sanh2021multitask} to encourage model generalization to unseen tasks in inference. It is an orthogonal concept with full fine-tuning and efficient fine-tuning, in other words, one can choose both full fine-tuning and efficient fine-tuning for instruction tuning. Instruction tuning is adopted in the graph domain for node classification \cite{tang2023graphgpt}, link prediction \cite{ye2023natural}, and graph-level tasks \cite{zhao2023gimlet}.
\end{itemize}

\noindent\textbf{Prompting} is a technique to apply language model for downstream task solving without updating the model parameters. 
One needs to formulate the test samples into natural language sequences and ask the language model to directly conduct inference based on the in-context demonstrations.
This is a technique particularly popular for large-scale autoregressive language models.
Apart from direct prompting, following-up works propose chain-of-thought prompting \cite{wei2022chain}, tree-of-thought prompting \cite{yao2023tree}, and graph-of-thought prompting \cite{besta2023graph}.

\begin{table*}[ht]
\caption{A collection of LLM reasoning methods on pure graph discussed in Section~\ref{sec:graph-reasoning}.
    We do not include the backbone models used in these methods studied in the original papers, as these methods generally apply to any LLMs.
    The ``Papers'' column lists the papers that study the specific methods.
}\label{tab:reasoning-methods}
\centering
\begin{adjustbox}{width=\textwidth}
    \begin{tabular}{p{0.17\textwidth}p{0.2\textwidth}p{0.4\textwidth}lp{0.15\textwidth}}
\toprule
    Method & Graph Format or Encoding & Reasoning Process & Reasoning Category & Papers \\
\midrule
    Zero-Shot
        & Verbalized edge or adjacency list.
        & Directly answering.
        & Direct Answering &~\cite{wang2023can, liu2023evaluating, guo2023gpt4graph, zhang2023llm4dyg, fatemi2023talk} \\
    \midrule

    Role Prompting
        & Verbalized edge or adjacency list.
        & Directly answering by designating a specific role to the LLM.\@
        & Direct Answering &~\cite{guo2023gpt4graph} \\
    \midrule

    Format Explanation
        & Verbalized edge or adjacency list.
        & Encouraging the LLM to explain the input graph format first.
        & Direct Answering &~\cite{guo2023gpt4graph} \\
    \midrule

    GraphLLM
        & Prefix tokens encoded by a graph encoder.
        & Directly answering.
        & Direct Answering &~\cite{chai2023graphllm} \\
    \midrule

    Few-Shot (In-Context Learning)
        & Verbalized edge or adjacency lists preceded with a few demonstrative examples.
        & Directly answering by following the examples.
        & Direct Answering &~\cite{wang2023can, liu2023evaluating, zhang2023llm4dyg, fatemi2023talk} \\
    \midrule

    Chain-of-Thought
        & Verbalized edge or adjacency lists preceded with a few demonstrative examples.
        & Reasoning through a series of intermediate reasoning steps in the generation following the examples.
        & Heuristic Reasoning &~\cite{wang2023can, liu2023evaluating, guo2023gpt4graph, zhang2023llm4dyg, fatemi2023talk, sun2023think} \\
    \midrule

    Self-Consistency
        & Verbalized edge or adjacency lists preceded with a few demonstrative examples.
        & Reasoning through a series of intermediate reasoning steps in generation, and then selecting the most consistent answer.
        & Heuristic Reasoning &~\cite{wang2023can} \\
    \midrule

    Build-a-Graph
        & Verbalized edge or adjacency list.
        & Reconstructing the graph in output, and then reasoning on the graph.
        & Heuristic Reasoning &~\cite{wang2023can, fatemi2023talk} \\
    \midrule

    Context-Summarization
        & Verbalized edge or adjacency list.
        & Directly answering by first summarizing the key elements in the graph.
        & Heuristic Reasoning &~\cite{guo2023gpt4graph} \\
    \midrule

    Reasoning-on-Graph
        & Retrieved paths from external graphs.
        & First, plan the reasoning process in the form of paths to be retrieved and then infer on the retrieved paths.
        & Heuristic Reasoning &~\cite{luo2023reasoning} \\
    \midrule

    Iterative Reading-then-Reasoning
        & Retrived neighboring edges or nodes from external graphs.
        & Iteratively retrieving neighboring edges or nodes from external graphs and inferring from the retrieved information.
        & Heuristic Reasoning &~\cite{jiang2023structgpt, sun2023think} \\
    \midrule

    Algorithmic Reasoning
        & Verbalized edge or adjacency list.
        & Simulating the reasoning process of a relevant algorithm in a generation.
        & Algorithmic Reasoning &~\cite{wang2023can} \\
    \midrule

    Calling APIs
        & External Knowledge Base.
        & Generate the reasoning process as (probably nested) API calls to be executed externally on the knowledge base.
        & Algorithmic Reasoning &~\cite{zhang2023graph, sun2023think} \\

\bottomrule
\end{tabular}
\end{adjustbox}
\end{table*}

\begin{table*}[ht]
\caption{A collection of pure graph reasoning problems studied in Section~\ref{sec:graph-reasoning}.
    $\mathcal{G} = (\mathcal{V}, \mathcal{E})$ denotes a graph with vertices $\mathcal{V}$ and edges $\mathcal{E}$. $v$ and $e$ denote individual vertices and edges, respectively.
    The ``Papers'' column lists the papers that study the problem using LLMs.
    The ``Complexity'' column lists the time complexity of standard algorithms for the problem, ignoring more advanced but complex algorithms that are not comparable to LLMs' reasoning processes.
}\label{tab:reasoning-problems}
\centering
\begin{adjustbox}{width=0.8\textwidth}
    \begin{tabular}{p{0.2\textwidth}p{0.3\textwidth}p{0.2\textwidth}p{0.2\textwidth}l}
\toprule
        \textbf{Problem} & \textbf{Definition} & \textbf{Applications} & \textbf{Typical Complexity} & \textbf{Papers} \\
\midrule
    Connectivity & Given a graph $\mathcal{G}$ and two nodes $u$ and $v$, tell if they are connected by a path.
        & Relationship Detection, Link Prediction
        & $O(|E|)$ or $O(V^2)$ &~\cite{wang2023can, liu2023evaluating} \\
    \midrule
    Neighbor Detection & Given a graph $\mathcal{G}$ and a node $v$, find the nodes connected to $v$.
        & Recommendation, Knowledge QA
        & $O(\min(|E|, |V|))$ &~\cite{guo2023gpt4graph} \\
    \midrule
    Node Degree & Given a graph $\mathcal{G}$ and a node $v$, find the number of edges connected to $v$.
        & Entity Popularity, Importance Ranking
        & $O(\min(|E|, |V|))$ &~\cite{liu2023evaluating, guo2023gpt4graph} \\
    \midrule
    Attribute Retrieval & Given a graph $\mathcal{G}$ with node-level information and a node $v$, return the attribute of $v$.
        & Recommendation, Node Classification, Node QA
        & $O(1)$ &~\cite{guo2023gpt4graph} \\
    \midrule
    Graph Size & Given a graph $\mathcal{G}$, find the number of nodes and edges.
        & Graph-level Classification
        & $O(|V| + |E|)$ &~\cite{guo2023gpt4graph} \\
    \midrule
    Cycle Detection & Given a graph $\mathcal{G}$, tell if it contains a cycle.
        & Loop Elimination, Program Loop Detection
        & $O(|V|)$ &~\cite{wang2023can} \\
    \midrule
    Diameter & Given a graph $\mathcal{G}$, find the diameter of $\mathcal{G}$.
        & Graph-level Classification, Clustering
        & $O(|V|^3)$ or $O(|V|^2\log|V| + |V||E|)$ &~\cite{guo2023gpt4graph} \\
    \midrule
    Topological Sort & Given a directed acyclic graph $\mathcal{G}$, find a topological ordering of its vertices so that for every edge $(u, v)$, $u$ comes before $v$ in the ordering.
        & Timeline Generation, Dependency Parsing, Scheduling
        & $O(|V| + |E|)$ &~\cite{wang2023can} \\
    \midrule
    Wedge or Triangle Detection & Given a graph $\mathcal{G}$ and a vertex $v$, identify if there is a wedge or triangle centered at $v$.
        & Relationship Detection, Link Prediction
        & $O(|V|+|E|)$ &~\cite{liu2023evaluating} \\
    \midrule
    Maximum Triplet Sum & Given a graph $\mathcal{G}$, find the maximum sum of the weights of three vertices that are connected.
        & Community Detection
        & $O(|V|^3)$ &~\cite{chai2023graphllm} \\
    \midrule
    Shortest Path & Given a graph $\mathcal{G}$ and two nodes $u$ and $v$, find the shortest path between $u$ and $v$.
        & Navigation, Planning
        & $O(|E|)$ or $O(V^2)$ &~\cite{wang2023can, liu2023evaluating, chai2023graphllm} \\
    \midrule
    Maximum Flow & Given a directed graph $\mathcal{G}$ with a source node $s$ and a sink node $t$, find the maximum flow from $s$ to $t$.
        & Transportation Planning, Network Design
        & $O(|V| |E|^2)$, $O(|E| |V| \log |V|)$ or $O(|V|^3)$ &~\cite{wang2023can} \\
    \midrule
    Bipartite Graph Matching & Given a bipartite graph $\mathcal{G}$ with two disjoint sets of vertices $\mathcal{V}_1$ and $\mathcal{V}_2$, find a matching between $\mathcal{V}_1$ and $\mathcal{V}_2$ that maximizes the number of matched pairs.
        & Recommendation, Resource Allocation, Scheduling
        & $O(|E|\sqrt{|V|})$ &~\cite{wang2023can, chai2023graphllm} \\
    \midrule
    Graph Neural Networks & Given a graph $\mathcal{G}$ with node features $\mathbf{X}$ of dimension $d$, simulate a graph neural networks with $l$ payers and return the encoded node features
        & Node Classification, Graph-level Classification
        & $O(ld|V|^2)$ &~\cite{wang2023can} \\
    \midrule
    Clustering Coefficient & Given a graph $\mathcal{G}$, find the clustering coefficient of $\mathcal{G}$.
        & Community Detection, Node Clustering
        & $O(|V|^3)$ &~\cite{guo2023gpt4graph} \\
    \midrule
    Substrcuture Counting & Given a graph $\mathcal{G}$ and a subgraph $\mathcal{G}'$, count the number of occurrences of $\mathcal{G}'$ in $\mathcal{G}$.
        & Pattern Matching, Subgraph Detection, Abnormality Detection
        & NP-Complete &~\cite{chai2023graphllm} \\
    \midrule
    Hamilton Path & Given a graph $\mathcal{G}$, find a path that visits every vertex exactly once.
        & Route Planning, Drilling Machine Planning, DNA Sequencing
        & NP-Complete &~\cite{wang2023can} \\
    \midrule
    (Knowledge) Graph QA & Given a (knowledge) graph $\mathcal{G}$ and a question $q$, find the answer to $q$.
        & Dialogue System, Smart Assistant, Recommendation
        & --- &~\cite{guo2023gpt4graph, luo2023reasoning, jiang2023structgpt, fatemi2023talk, sun2023think} \\
    \midrule
    Graph Query Language Generation & Given a graph $\mathcal{G}$ and a query $q$, generate a query language that can be used to query $\mathcal{G}$.
        & Graph Summarization, FAQ Generation, Query Suggestions
        & --- &~\cite{guo2023gpt4graph} \\
    \midrule
    Node Classification & Given a graph $\mathcal{G}$, predict the class of a node $v$.
        & Recommendation, User Profiling, Abnormality Detection
        & --- &~\cite{guo2023gpt4graph, zhang2023graph} \\
    \midrule
    Graph Classification & Given a graph $\mathcal{G}$, predict the class of $\mathcal{G}$.
        & Molecule Property Prediction, Moledule QA, Graph QA
        & --- &~\cite{guo2023gpt4graph, zhang2023graph} \\
\bottomrule
\end{tabular}
\end{adjustbox}
\end{table*}

\begin{table*}[htb]
\caption{Summary of large language models on text-attributed graphs. Role of LM: ``TE'', ``SE'', ``ANN'' and ``AUG'' denote text encoder, structure encoder, annotator (labeling the node/edges), and augmentator (conduct data augmentation). Task: ``NC'', ``UAP'', ``LP'', ``Rec'', ``QA'', ``NLU'', ``EC'', ``LM'', ``RG'' denote node classification, user activity prediction, link prediction, recommendation, question answering, natural language understanding, edge classification, language modeling, and regression task.}\label{tab:node-text-papers}
\centering
\begin{adjustbox}{width=0.8\textwidth}
\begin{tabular}{lllllll}
\toprule
Approach & Category & Role of LM & LM Size & Focus & Task \\ \midrule
GNN-LM \cite{meng2021gnn}  & LLM as Encoder & TE & 237M & Task & LM  \\ \midrule
GIANT \cite{chien2021node} & LLM as Encoder & TE & 110M & Task & NC  \\ \midrule
TextGNN \cite{zhu2021textgnn}  & LLM as Encoder & TE & 110M & Task & Search  \\ \midrule
AdsGNN \cite{li2021adsgnn}  & LLM as Encoder & TE & 110M & Task & Search  \\ \midrule
LM-GNN \cite{ioannidis2022efficient}  & LLM as Encoder & TE & 110M & Efficiency & NC, LP, EC \\ \midrule
GraD \cite{mavromatis2023train}  & LLM as Encoder & TE & 110M/66M & Efficiency & LP, NC  \\ \midrule
TAPE \cite{he2023explanations}  & LLM as Encoder & TE, AUG & 129M/GPT-3.5 & Task & NC  \\ \midrule
SimTeG \cite{duan2023simteg}  & LLM as Encoder & TE & 80M/355M & Task & NC, LP  \\ \midrule
LLM-GNN \cite{chen2023label}  & LLM as Encoder & ANN & GPT-3.5 & Task & NC  \\ \midrule
ENG \cite{yu2023empower} & LLM as Encoder & TE, AUG & 80M/GPT-3.5 & Task & NC  \\ \midrule
SPECTER \cite{cohan2020specter}  & LLM as Predictor & TE & 110M & Representation & NC, UAP, LP, Rec  \\ \midrule
GraphFormers \cite{yang2021graphformers} & LLM as Predictor & TE, SE & 110M & Representation & LP    \\ \midrule
GreaseLM \cite{zhang2022greaselm}  & LLM as Predictor & TE, SE & 355M & Task & QA  \\ \midrule
SciNCL \cite{ostendorff2022neighborhood}  & LLM as Predictor & TE & 110M & Representation & NC, UAP, LP, Rec  \\ \midrule
MICoL \cite{zhang2022metadata}  & LLM as Predictor & TE & 110M & Supervision & NC  \\ \midrule
LinkBERT \cite{yasunaga2022linkbert}  & LLM as Predictor & TE & 110M & Pretraining & QA, NLU \\ \midrule
Heterformer \cite{jin2023heterformer}  & LLM as Predictor & TE, SE & 110M & Representation & NC, LP  \\ \midrule
E2EG \cite{dinh2022e2eg} & LLM as Predictor & TE & 66M & Task & NC  \\ \midrule
TwHIN-BERT \cite{zhang2023twhin}  & LLM as Predictor & TE & 110M/355M & Pretraining & NC, LP  \\ \midrule
Edgeformers \cite{jin2023edgeformers}  & LLM as Predictor & TE, SE & 110M & Representation & NC, LP, EC   \\ \midrule
Patton \cite{jin2023patton}  & LLM as Predictor & TE, RE & 110M & Pretraining & NC, LP, Search   \\ \midrule
InstructGLM \cite{ye2023natural} & LLM as Predictor & TE, SE & 250M/7B & Generalization & NC, LP  \\ \midrule
GNP \cite{tian2023graph}  & LLM as Predictor & TE, SE & 3B/11B & Task & QA  \\ \midrule
Touchup-G \cite{zhu2023touchup} & LLM as Predictor & TE & 110M & Representation & NC, LP  \\ \midrule
DGTL \cite{qin2023disentangled} & LLM as Predictor & TE, SE & 13B & Task & NC    \\ \midrule
GraphText \cite{zhao2023graphtext} & LLM as Predictor & TE, SE & GPT-3.5/4 & Task & NC  \\ \midrule
GraphGPT \cite{tang2023graphgpt} & LLM as Predictor & TE, SE & 7B & Generalization & NC  \\ \midrule
METERN \cite{jin2023learning} & LLM as Predictor & TE, RE & 110M & Representation & NC, LP, Rec, RG    \\ \midrule
LTRN \cite{zhang2021minimally} & LLM as Aligner & TE & 110M & Supervision & NC  \\ \midrule
GLEM \cite{zhao2022learning} & LLM as Aligner & TE & 110M & Task & NC  \\ \midrule
G2P2 \cite{wen2023augmenting} & LLM as Aligner & TE & 110M & Supervision & NC  \\ \midrule
ConGraT \cite{brannon2023congrat} & LLM as Aligner & TE & 110M/82M & Representation & LP, LM, NC  \\ \midrule
GRENADE \cite{li2023grenade} & LLM as Aligner & TE & 110M & Representation & NC, LP  \\ \midrule
THLM \cite{zou2023pretraining} & LLM as Aligner & TE & 110B & Pretraining & NC, LP   \\ 
\bottomrule
\end{tabular}
\end{adjustbox}
\end{table*}

\begin{table*}[htb]
\caption{A summarization of Graph-Aware LLM finetuning objectives on text-attributed graphs. $v^+_i$ and $v^-_i$ denote a positive training node and a negative training node to $v_i$, respectively.}\label{tab:graph-finetune-objective}
\centering
\begin{adjustbox}{width=0.90\textwidth}
\begin{tabular}{lccc}
\toprule
Method & positive $v^+_i$ & negative $v^-_i$  & Objective $f(\cdot)$ \\ \midrule
\multirow{2}{*}{SPECTER \cite{cohan2020specter}} & \multirow{2}{*}{$(v_i, v^+_i)\in \mathcal{E}$} & $(v_i, v^-_i)\notin \mathcal{E}$; & \multirow{2}{*}{$\max\{||\mathbf{h}_{v_i}-\mathbf{h}_{v^+_i}||_2 - ||\mathbf{h}_{v_i}-\mathbf{h}_{v^-_i}||_2 + m, 0\}$} \\
 & & $(v_i, v_u)\in \mathcal{E}$, $(v_u, v^-_i)\in \mathcal{E}$, $(v_i, v^-_i)\notin \mathcal{E}$ & \\ \midrule
SciNCL \cite{ostendorff2022neighborhood} & $||\mathbf{h}_{v_i}-\mathbf{h}_{v^+_i}||_2\in (k^+-c^+;k^+]$ & $||\mathbf{h}_{v_i}-\mathbf{h}_{v^-_i}||_2\in (k^-_{hard}-c^-_{hard};k^-_{hard}]$  & $\max\{||\mathbf{h}_{v_i}-\mathbf{h}_{v^+_i}||_2 - ||\mathbf{h}_{v_i}-\mathbf{h}_{v^-_i}||_2 + m, 0\}$ \\ \midrule
Touchup-G \cite{zhu2023touchup} & $(v_i, v^+_i)\in \mathcal{E}$ & $(v_i, v^-_i)\notin \mathcal{E}$ & $\text{log}(\mathbf{h}_{v_i} \cdot \mathbf{h}_{v^+_i}) + \text{log}(1-\mathbf{h}_{v_i} \cdot \mathbf{h}_{v^-_i})$  \\ \midrule 
TwHIN-BERT \cite{zhang2023twhin} & $\cos(\mathbf{x}_{v_i},\mathbf{x}_{v^+_i})<k$ & in-batch random & $-\log\frac{\exp(\cos(\mathbf{h}_{v_i}, \mathbf{h}_{v^+_i})/ \eta)}{\sum_{v^-_i}\exp(\cos(\mathbf{h}_{v_i}, \mathbf{h}_{v^-_i})/ \eta)}$    \\ \midrule 
MICoL \cite{zhang2022metadata}  &  $v^+_i\in N_M(v_i)$  & in-batch random & $-\log\frac{\exp(\cos(\mathbf{h}_{v_i}, \mathbf{h}_{v^+_i})/ \eta)}{\sum_{v^-_i}\exp(\cos(\mathbf{h}_{v_i}, \mathbf{h}_{v^-_i})/ \eta)}$   \\ \midrule 
E2EG \cite{dinh2022e2eg} & $(v_i, v^+_i)\in \mathcal{E}$ & $(v_i, v^-_i)\notin \mathcal{E}$ & $\text{log}(\mathbf{h}_{v_i} \cdot \mathbf{h}_{v^+_i}) + \sum_{v^-_i} \text{log}(1-\mathbf{h}_{v_i} \cdot \mathbf{h}_{v^-_i})$  \\  
\bottomrule
\end{tabular}
\end{adjustbox}
\end{table*}

\begin{figure*}[!t]
\centering
\includegraphics[width=16cm]{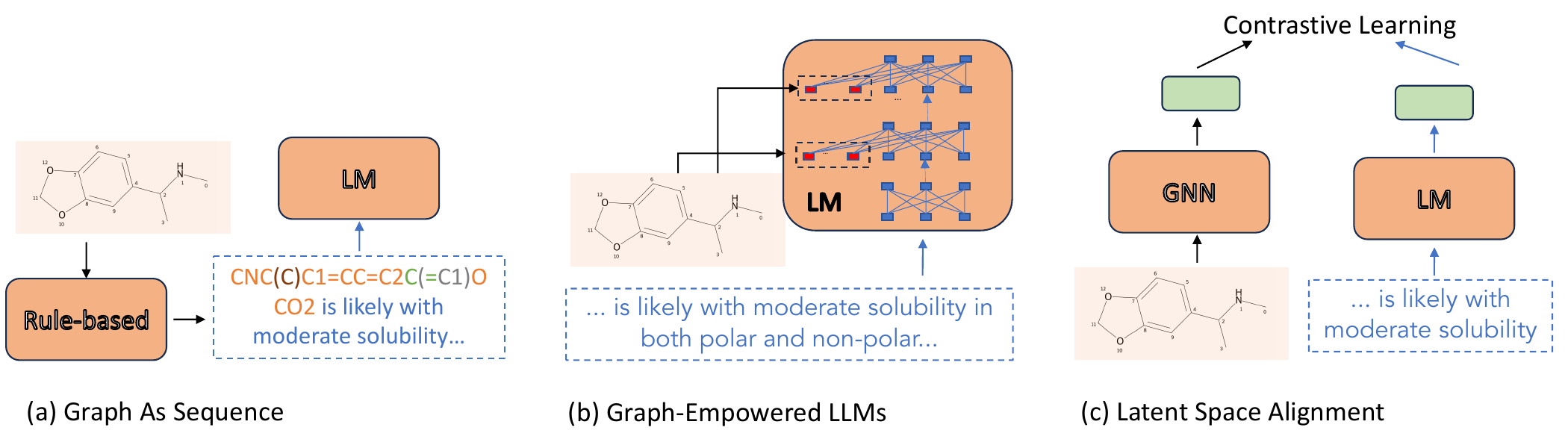}
\caption{The illustration covers LLM methods for text-paired graphs, including two LLM predictor methods: (a) Rule-based Linearization, (b) Graph-Empowered LMs, and one LLM aligner method: (c) Latent Space Alignment. }\label{fig::text-graph-pair-methods}
\end{figure*}

\begin{table*}[htb]
    \centering
    \caption{Model collection in Section~\ref{sec:graph-text} for text-paired graphs. ``Lin.'' and ``Vec.'' represent Linearized Graph Encoding and Vectorized Graph Encoding. ``Classif.'', ``Regr.'', ``NER'', ``RE'', ``Retr.'', ``Gen.'', ``Cap.'' represent classification, regression, named entity recognition, relation extraction, (molecule) graph retrieval, (molecule) graph generation, (molecule) graph captioning.}
    \label{tab:graph-text-model-v2}
    \begin{adjustbox}{width=0.8\textwidth}
    \begin{tabular}{lllllll}
    \toprule
        & Model  & LM Encoder & Graph Encoder & Gen. Decoder & LM Size & Task\\
        \midrule
        \multirow{30}{*}{\rotatebox{90}{LLM as Predictor in Section~\ref{subsec:graph-text-LM-Centric}}}
        & SMILES-BERT~\cite{smilesbert} & Transformer~\cite{vaswani2017attention} & Linearized & N.A. & 30M-565M & Classification \\
        \cmidrule(lr){2-7}
        & Text2Mol~\cite{edwards2021text2mol}  & SciBERT~\cite{beltagy2019scibert} & GCN & Transformer~\cite{vaswani2017attention}  & $\geq$110M & Retrieval \\
        \cmidrule(lr){2-7}
        & MolGPT~\cite{molgpt}  & N.A. & Linearized & GPT & 6M & Generation \\
        \cmidrule(lr){2-7}
        & Chemformer~\cite{chemformer}  & BART~\cite{lewis2019bart} & Linearized & BART~\cite{lewis2019bart} & 45M-230M & Regression, Gen. \\
        \cmidrule(lr){2-7}
        & \multirow{2}{*}{KV-PLM~\cite{kvplm}} & \multirow{2}{*}{BERT~\cite{devlin2018bert}} & \multirow{2}{*}{Linearized} & \multirow{2}{*}{N.A} & \multirow{2}{*}{110M-340M} & Classif., NER, \\
        & & & & & & RE, Retrieval \\
        \cmidrule(lr){2-7}
        & MFBERT~\cite{mfbert}  & RoBERTa~\cite{liu2019roberta} & Linearized & N.A. & 110M-340M & Classification \\
        \cmidrule(lr){2-7}
        & Galatica~\cite{galatica} & N.A. & Linearized & Transformer~\cite{vaswani2017attention} & 125M-120B & Classification \\
        \cmidrule(lr){2-7}
        & MolT5~\cite{edwards2022translation}  & T5.1.1~\cite{raffel2020exploring} & Linearized & Transformer & 80M-780M & Gen., Cap. \\
        \cmidrule(lr){2-7}
        & \multirow{2}{*}{Text+Chem T5~\cite{tct}} & \multirow{2}{*}{T5~\cite{raffel2020exploring}} & \multirow{2}{*}{Linearized} & \multirow{2}{*}{T5~\cite{raffel2020exploring}} & \multirow{2}{*}{80M-780M} & Classif, Gen., \\
        & & & & & & Caption \\
        \cmidrule(lr){2-7}
        & \multirow{3}{*}{LLM-ICL~\cite{llmicl}}  & \multirow{3}{*}{N.A.} & \multirow{3}{*}{Linearized} & GPT-3.5/4 & \multirow{3}{*}{$\geq$ 780M} & Classification \\
        & & & & LLaMA2~\cite{touvron2023llama} & & Generation \\
        & & & & Galactica~\cite{galatica} & & Caption \\
        \cmidrule(lr){2-7}
        & GIMLET~\cite{zhao2023gimlet}  & T5~\cite{raffel2020exploring} & GT & T5~\cite{raffel2020exploring} & 80M-780M & Classif., Regr. \\
        \cmidrule(lr){2-7}
        & \multirow{2}{*}{MolXPT~\cite{molxpt}} & \multirow{2}{*}{N.A.} & \multirow{2}{*}{Linearized} & \multirow{2}{*}{GPT-2} & \multirow{2}{*}{350M} & Classif, Gen., \\
        & & & & & & Caption \\
        \cmidrule(lr){2-7}
        & ChatMol~\cite{chatmol} & T5~\cite{raffel2020exploring} & Linearized & T5~\cite{raffel2020exploring} & 80M-780M & Gen., Cap. \\
        \cmidrule(lr){2-7}
        & MolReGPT~\cite{molregpt} & N.A. & Linearized & GPT-3.5 & N.A. & Gen., Cap.\\
        \cmidrule(lr){2-7}
        & RT~\cite{regt} & N.A. & Linearized & XLNet~\cite{yang2019xlnet} & 27M & Regr. Gen. \\
        \cmidrule(lr){2-7}
        & LLM4Mol~\cite{llm4mol} & RoBERTa~\cite{liu2019roberta} & Linearized & GPT-3.5 & N.A. & Classif, Regr. \\
        \cmidrule(lr){2-7}
        & LLaMA-Mol~\cite{llamamol} & N.A. & Linearized & LLaMA~\cite{touvron2023llama} & 7B & Regr., Gene. \\
        \cmidrule(lr){2-7}
        & Prot2Text~\cite{prot2text} & N.A. & GNN & GPT-2 & 256M-760M & Caption \\
        \cmidrule(lr){2-7}
        & CatBERTa~\cite{catberta} & N.A. & Linearized & RoBERTa~\cite{liu2019roberta} & N.A. & Regression \\
        \cmidrule(lr){2-7}
        & ReLM~\cite{relm}  & N.A. & GNN & GPT-3.5 & N.A. & Classification \\
        \midrule
        \multirow{15}{*}{\rotatebox{90}{LLM as Aligner in Section ~\ref{subsec:graph-text-LM-GM}}}
        & \multirow{2}{*}{MoMu~\cite{momu}}  & \multirow{1}{*}{SciBERT~\cite{beltagy2019scibert}} & \multirow{2}{*}{GNN} & \multirow{1}{*}{MolT5~\cite{edwards2022translation}} & \multirow{2}{*}{82M-782M} & Classif, Gen., \\
        & & KV-PLM~\cite{kvplm} & & MoFlow~\cite{moflow} & & Caption, Retr. \\
        \cmidrule(lr){2-7}
        & \multirow{2}{*}{MoleculeSTM~\cite{moleculestm}}  & \multirow{2}{*}{BART~\cite{lewis2019bart}} & GIN & \multirow{2}{*}{BART~\cite{lewis2019bart}} & \multirow{2}{*}{45M-230M} & Classif, Gen., \\
        & & & Linearized & & & Caption \\
        \cmidrule(lr){2-7}
        & \multirow{2}{*}{CLAMP~\cite{clamp}} & BioBERT~\cite{biobert} & GNN & \multirow{2}{*}{N.A.} & \multirow{2}{*}{$\leq$11B} & Classification \\
        & & CLIP~\cite{radford2021learning}, T5~\cite{raffel2020exploring} & Lin., Vec. & & & Retrieval \\
        \cmidrule(lr){2-7}
        & \multirow{2}{*}{MolFM~\cite{molfm}} & \multirow{2}{*}{BERT~\cite{devlin2018bert}} & \multirow{2}{*}{GIN} & \multirow{2}{*}{MolT5~\cite{edwards2022translation}} & \multirow{2}{*}{61.8M} & Classif., Gen. \\
         & & & & & & Caption, Retr. \\
        \cmidrule(lr){2-7}
        & MoMu-v2~\cite{momuv2}  & SciBERT~\cite{beltagy2019scibert} & GIN & N.A. & 82M-782M & Classification \\
        \cmidrule(lr){2-7}
        & \multirow{2}{*}{GIT-Mol~\cite{gitmol}} & \multirow{2}{*}{SciBERT~\cite{beltagy2019scibert}} & GIN & \multirow{2}{*}{MolT5~\cite{edwards2022translation}} & \multirow{2}{*}{190M-890M} & Classif, Gen.\\
        & & & Linearized & & & Caption \\
        \cmidrule(lr){2-7}
        & \multirow{2}{*}{MolCA~\cite{molca}} & \multirow{2}{*}{Galactica~\cite{galatica}} & \multirow{2}{*}{GIN} & \multirow{2}{*}{N.A.} & \multirow{2}{*}{100M-877M} & Classif., Regr.,\\
        & & & & & & Retrieval \\
    \bottomrule
    \end{tabular}
    \end{adjustbox}
\end{table*}


\end{document}